\documentclass[11pt, letter]{article}

\usepackage[margin=1.1in]{geometry} 
\usepackage{parskip}
\usepackage{longtable,color,natbib}
\usepackage{booktabs} 
\usepackage{array} 
\usepackage{paralist} 
\usepackage{verbatim} 
\usepackage{subfigure} 
\usepackage{appendix}
\usepackage{amssymb}
\usepackage{algorithm,algorithmic}
\usepackage{amsmath}
\usepackage{bbm}

\definecolor{linkcolor}{rgb}{0.1,0,0.7}
\definecolor{urlcolor}{rgb}{1,0,0}
\usepackage[unicode=true,colorlinks,citecolor=linkcolor,linkcolor=linkcolor,urlcolor=blue]{hyperref}
\usepackage{graphicx} 
\usepackage{epstopdf}
\usepackage{bm}

\newtheorem{Theorem}{Theorem}[section]
\newtheorem{Definition}[Theorem]{Definition}

\newtheorem{Lemma}[Theorem]{Lemma}

\newtheorem{Remark}[Theorem]{Remark}

\newtheorem{Assumption}[Theorem]{Assumption}

\numberwithin{equation}{section}
\usepackage{hhline}
\usepackage{verbatim}
\usepackage{eurosym}

\newcommand{\nc}{\newcommand}
\nc{\ind}{\mathds{1}}
\def \trans{^{\scriptscriptstyle{\intercal}}}

\newcommand{\R}{\mathbb{R}}

\newcommand{\E}{\mathcal{E}}
\newcommand{\F}{\mathcal{F}}

\allowdisplaybreaks[4]

\DeclareMathOperator{\esssup}{esssup}

\def\esssup_#1{\underset{#1}{\mathrm{ess\,sup\, }}}
\def\essinf_#1{\underset{#1}{\mathrm{ess\,inf\, }}}
\def\argmax_#1{\underset{#1}{\mathrm{arg\,max\, }}}
\def\argmin_#1{\underset{#1}{\mathrm{arg\,min\, }}}

\def\b1{\bf 1}

\def \N{\mathbb{N}}
\def \R{\mathbb{R}}

\def \E{\mathbb{E}}
\def \F{\mathbb{F}}

\def \P{\mathbb{P}}

\def \Ac{{\cal A}}

\def \Cc{{\cal C}}

\def \Fc{{\cal F}}
\def \Gc{{\cal G}}
\def \Hc{{\cal H}}

\def \Pc{{\cal P}}

 \def \Nc{{\cal N}}

\def \Uc{{\cal U}}

\def \Wc{{\cal W}}

\def \Xc{{\cal X}}

\def\eqref#1{{\rm(\ref{#1})}}
\def\beqs{\begin{eqnarray*}}
\def\enqs{\end{eqnarray*}}
\def\beq{\begin{eqnarray}}
\def\enq{\end{eqnarray}}

\begin{document}

%
%

\title{Continuous time q-learning for mean-field control problems}

\author{Xiaoli Wei \thanks{Email: xiaoli.wei@hit.edu.cn, Institute for Advance Study in Mathematics, Harbin Institute of Technology, China.}
\and
Xiang Yu \thanks{Email: xiang.yu@polyu.edu.hk, Department of Applied Mathematics, The Hong Kong Polytechnic University, Kowloon, Hong Kong.}
}
\date{\vspace{-0.6cm}}

\maketitle
\begin{abstract}
This paper studies the q-learning, recently coined as the continuous time counterpart of Q-learning by \cite{jiazhou2022}, for continuous time mean-field control problems in the setting of entropy-regularized reinforcement learning. In contrast to the single agent's control problem in \cite{jiazhou2022}, we reveal that two different q-functions naturally arise in mean-field control problems: (i) the integrated q-function (denoted by $q$) as the first-order approximation of the integrated Q-function introduced in \cite{GGWX23}, which can be learnt by a weak martingale condition using all test policies; and (ii) the essential q-function (denoted by $q_e$) that is employed in the policy improvement iterations. We show that two q-functions are related via an integral representation. Based on the weak martingale condition and our proposed searching method of test policies, some model-free learning algorithms are devised. In two examples, one in LQ control framework and one beyond LQ control framework, we can obtain the exact parameterization of the optimal value function and q-functions and illustrate our algorithms with simulation experiments.
\end{abstract}
\vspace{0.1in}

\section{Introduction}

Mean-field control (MFC) problems, also known as McKean-Vlasov control problems, concern stochastic systems of large population where the agents interact through the distribution of their states and the optimal decision is made by a social planner to attain the Pareto optimality. It has been seen rapid progress in both theories and applications in the study of MFC problems in the past two decades. The comprehensive survey of this topic and related studies can be found in \cite{CarD} and \cite{CarD2}.

By nature, the model parameters in the stochastic system with a large population of agents are generally difficult to observe or estimate, and it is a desirable research direction to design efficient learning algorithms for solving the MFC problems under the unknown environment. Reinforcement learning (RL) provides many well-suited learning algorithms for this purpose that enables the agent to learn the optimal control through the trial-and-error procedure. Through the exploration and exploitation in RL, the agent takes an action and observes a reward outcome that signals the influence of the agent's action such that the agent can learn to select actions based on past experiences and by making new choices. Recently, many exploration-exploitation RL algorithms for single agent's control problems were quickly adopted and generalized in many multi-agent and mean-field reinforcement learning applications.

Despite its substantial success in wide applications, the theoretical study of reinforcement learning has been predominantly limited to discrete time models, especially in the MFC and mean-field game problems. \cite{GGWX23} discussed the correct form of Q-function in an integral form in Q-learning and established the dynamic programming principle (DPP) for learning MFC problems. \cite{GGWX21a, GGWX21b} proposed a model-free Q learning in a centralized way and an actor-critic algorithm in a decentralized manner for MFC problems with convergence analyses. \cite{CLT23} studied MFC with common noise under both closed-loop and open-loop policies through the lens of mean-field Markov decision process, which also unveiled the correct form of the Q-function and its dynamic programming principle and adapted existing (deep) RL methods to the mean-field setting. \cite{AFL2022, AFHR2023} studied  unified two-timescale RL algorithms to solve infinite horizon asymptotic MFC and mean-field game problems in finite and continuous state-action space, respectively. The corresponding convergence result is studied in the subsequent work \cite{AFLZ23}. There also exists a variety of model-based or model-free RL algorithms for MFCs that can be found, for instance in  \cite{Mondaletal2022, Mondaletal2023, Pasztoretal2021}. On the other hand, for continuous time stochastic control problems by a single agent, \cite{Wangetal2021}, \cite{JZ22a, JZ22b, jiazhou2022} have laid the overarching theoretical foundation for reinforcement learning in continuous time with continuous state space and possibly continuous action space. In particular, \cite{jiazhou2022} developed a continuous time q-learning theory by considering the first order approximation of the conventional Q-function. Given a stochastic policy, the value function and the associated q-function can be characterized by martingale conditions of some stochastic processes in \cite{jiazhou2022} in both on-policy and off-policy settings. Several actor-critic algorithms are devised therein for solving underlying RL problem. Convergence or regret analyses of continuous-time single-agent stochastic control problems have been conducted in \cite{giegrichetal2024, STZ2021} for linear-quadratic or linear-convex models. For continuous time mean-field LQ games, \cite{GuoXZ} examined the theoretical justification that entropy regularization helps stabilizing and accelerating the convergence to the Nash equilibrium. \cite{LiLiXu} concerned the model-based policy iteration reinforcement learning method for continuous time infinite horizon mean-field LQ control problems. Recently, \cite{FGLPS23} generalized the policy gradient algorithm in \cite{JZ22b} to continuous time MFC problems and devised actor-critic algorithms based on a gradient expectation representation of the value function, where the value function and the policy are learnt alternatively via the observed samples of the state and a model-free estimation of the population distribution.

As demonstrated in \cite{jiazhou2022}, one important advantage of continuous time framework lies in its robustness with respect to the time discretization in the algorithm design and implementations. Inspired by \cite{jiazhou2022}, we are particularly interested in whether, and if yes, how the continuous time q-learning can be applied in learning McKean-Vlasov control problems in the mean-field model with infinitely many interacting agents. Contrary to \cite{jiazhou2022}, our first contribution is to reveal that two distinct q-functions are generically needed to learn continuous time McKean-Vlasov control problems. In fact, due to the feature of mean-field interactions with the population, it has been shown in \cite{GGWX23, CLT23} in the discrete time setting that the time consistency and DPP only hold for the \textit{so-called} integrated Q-function (or IQ) for learning MFC problems, where the IQ-function depends on the distribution of both the state and the control, i.e., the state space of the state variable and the action variable needs to be appropriately enlarged to the Wasserstein space of probability measures so that the time-consistency can be guaranteed. Similar to the counterpart in \cite{GGWX23, CLT23}, we also show that the correct definition of the continuous time q-function, as the first order approximation of the IQ-function, needs to be defined on the distribution of the state and the control as an integral form of the Hamiltonian operator. This integral form of the q-function is called the integrated q-function in the present paper, which is crucial in establishing a weak martingale characterization of the value function and the q-function using all test policies in a neighbourhood of the target policy. However, on the other hand, from the relaxed control formulation with the entropy regularizer and its associated exploratory HJB equation, the optimal policy can be obtained as a Gibbs measure related to the Hamiltonian operator directly. Therefore, the integrated q-function actually can not be utilized directly to learn the optimal policy. Instead, a proper way is to introduce another q-function that is defined as the Hamiltonian (without the integral form) plus the temporal dispersion, which can be employed in the policy improvement iterations. To distinguish two different q-functions in our setting, we shall call the first order approximation of the IQ-function as the continuous time integrated q-function, denoted by $q$ (see Definition \ref{def:coupled-q-function}); and we shall name the second q-function related to the policy improvement as the \textit{essential} q-function, denoted by $q_e$ (see Definition \ref{defqe}).

One key finding of the present paper is the integral representation between $q$ and $q_e$ involving all test policies, see the relationship in \eqref{relationq_dq_cbarq}. As a result, we can employ the weak martingale characterization of the integrated q-function (see Theorems \ref{thm:unknown-q-function} and \ref{thm:unknown-coupled-optimal-q}) to devise the learning algorithm in the following order: $(i)$ We first parameterize the value function $J^{\theta}(t,\mu)$ and the essential q-function $q_e^{\psi}$, and then obtain the parameterized q-function $q^{\psi}$ from the integral relationship that shares the same parameter $\psi$ of $q_e^{\psi}$. $(ii)$ By minimizing the weak martingale loss under all test policies, we devise updating rules for the parameters in the integrated $q$-function $q^{\psi}$. Here, a new challenge we encounter is to decide the searching rule of test policies in the weak martingale characterization. To address this issue, we consider the minimization of the weak martingale loss robust with respect to all test policies, i.e., the updating rule is determined by a minmax problem. We propose a method based on the average of some test policies close to the target policy, which does not create new parameters. In addition, it is worth noting that the test policy instead of the target policy will be used to generate samples and observations due to the dependence of the distribution of the action in the integrated q-function, therefore making our learning algorithms similar to the off-policy learning in the conventional Q-learning, see Remark \ref{testpolicyexp}. $(iii)$ After the updating of the parameter $\psi$ in $q$ from step $(ii)$, the parameter of $q_e$ is also updated that gives the updating rule of the associated policy.

On the other hand, it is shown in \cite{GGWX23} that the IQ-function in the discrete time framework is generally nonlinear in the distribution of control, and hence it is difficult to characterize the distribution of the optimal policy. By contrast, it is interesting to see that our integrated q-function, as the first order derivative of the IQ-function with respect to time, admits a linear integral form of the distribution in \eqref{relationq_dq_cbarq}, which is more suitable for the name of ``integrated function" of the distribution of the population and allows the explicit connection between the optimal policy and the essential q-function, see Remark \ref{continuousRM}.

To illustrate our model-free q-learning algorithms, we study two concrete examples in financial applications, namely the optimal mean-variance portfolio problem in the LQ control framework and the mean-field optimal R$\&$D investment and consumption problem beyond the LQ control framework. In both examples, we can derive the explicit expressions of the optimal value function and two q-functions from the exploratory HJB equation and hence obtain their exact parameterized forms. From simulation experiments, we illustrate the satisfactory performance of our q-learning algorithm and the proposed searching method of test policies.

The remainder of this paper is organized as follows. Section \ref{sec:formulation} introduces the exploratory learning formulation for continuous time McKean-Vlasov control problems. In Section \ref{sec:q-function}, the definitions of the continuous time integrated q-function and the essential q-function are given and their relationship is established. The weak martingale characterization of the value function and two q-functions using test policies are established in Section \ref{sec:martingale_characterization}. Some offline and online q-learning algorithms using the average-based test policies are presented in Section \ref{sec:algorithms}. Section \ref{sec:application} examines two concrete financial applications and numerically illustrates our q-learning algorithms in some simulation experiments. Section \ref{sec:conclusion} concludes the contributions of the paper and discusses some directions for future research.

\section{Problem Formulation}\label{sec:formulation}

\subsection{Strong Control Formulation}
Let $(\Omega, \Fc, \P)$ be the probability space that supports a $n$-dimensional Brownian motion $W=(W_s)_{s \in [0, T]}$. We denote by $\F^W = (\Fc_s^W)_{s \in [0,T]}$ the $\P$-completion filtration of $W$. We assume that there is a sub-algebra $\Gc$ of $\Fc$ such that $\Gc$ is independent of $\F^W$ and  is ``rich enough" in the sense that for any $\mu \in \Pc_2(\R^d)$, there exists a $\Gc$-measurable random variable $\xi$ such that $\P_\xi = \mu$.  Indeed, such $\Gc$ exists if and only if there exists a $\Gc$-measurable random variable $U$ with the uniform distribution $\Uc([0, 1])$ that is independent of $W = (W_s)_{s \geq 0}$. We denote by $\F = (\Fc_s)_{s \geq 0}$ the filtration defined by $\Fc_s = \Fc_s^W \vee \Gc$, see section 2 in \cite{cossoetal2020} for more details.

The state dynamics of the controlled McKean-Vlasov SDE is given by
\begin{align}
dX_s = b(s, X_s, \P_{X_s}, a_s) ds  + \sigma(s, X_s, \P_{X_s}, a_s) dW_s,\;\;\; X_t = \xi \sim \mu, \label{Xdy}
\end{align}
where $\P_{X_s}$ denotes the probability distribution of $X_s$. In the present paper, we only consider the model without common noise, leaving the case with common noise as future study.

The goal of the McKean-Vlasov control problem is to find the optimal $\F$-progressively measurable sequence of actions $\{a_s\}_{t \leq s \leq T}$ valued in the space $\Ac\subset \R^m$ that maximizes the expected discounted total reward
\begin{align}\label{equ:strong-value}
V^*(t,\mu):=\sup_{\{a_s\}_{t\leq s\leq T}} \E\left[\int_t^T e^{-\beta(s-t)} r(s, X_s, \P_{X_s}, a_s)ds + e^{-\beta (T -t)}g( X_T, \P_{X_T})\Big| \P_{X_t}=\mu\right].
\end{align}
One can show that the optimal value function is law-invariant and $V^*: [0, T] \times \Pc_2(\R^d) \to \R$ satisfies the dynamic programming principle
\begin{align*}
V^*(t, \mu) = \sup_{\{a_s\}_{t\leq s\leq T}} \E\left[\int_t^{t+h} e^{-\beta(s-t)} r(s, X_s, \P_{X_s}, a_s)ds + e^{-\beta h} V^*(t+h, \P_{X_{t+h}})\right].
\end{align*}
Furthermore, when $b$, $\sigma$, $r$ and $g$ are known, the optimal value function satisfies the following dynamic programming equation
\begin{align*}
&\frac{\partial V^*}{\partial t}(t, \mu) - \beta V^*(t, \mu) + \int_{\R^d} \sup_{a \in \Ac}H\big(t, x, \mu, a, \partial_\mu V^*(t, \mu)(x), \partial_x\partial_\mu V^*(t, \mu)(x)\big)\mu(dx) = 0,
\end{align*}
with the terminal condition $V^*(T, \mu) = \int_{\R^d} g(x, \mu)\mu(dx): = \hat g(\mu)$, where $\partial_\mu V^*(t, \mu)(x)$ is defined via the lifting identification and it is called L-derivative of $V^*$ with respect to the measure $\mu$, see \cite{Lions} and Definition 5.22 in  \cite{CarD}, and $\partial_x\partial_\mu V^*(t, \mu)(x)$ denotes the partial derivative of $\partial_\mu V^*(t, \mu)(x)$ with respect to $x$, and the Hamiltonian operator $H$ is defined by
\begin{align}\label{operatorH}
H(t, x, \mu, a, p, q) & =  b(t, x, \mu, a)\trans p + \frac{1}{2} {\rm Tr}\big(\sigma\sigma\trans(t, x, \mu, a)
q\big) + r(t, x, \mu, a).
\end{align}

\subsection{Exploratory Formulation} We now consider the situation when the model is unknown, i.e., we do not have the exact information of the model parameters $b$ and $\sigma$ in state dynamics \eqref{Xdy} and the reward function $r$ in \eqref{equ:strong-value}. To determine the optimal control in face of the unknown model, we choose to apply the reinforcement learning based on the trial-and-error procedure. That is, what the representative agent (social planner) can do is to try a sequence of actions $(a_s)_{s\in[0,T]}$ and observe the corresponding state process $(X_s^a)_{s\in [0,T]}$ and the distribution $(\P_{X^a_s})_{s\in[0,T]}$ along with a stream of discounted running rewards $r(s, X^a_s, \P_{X^a_s}, a_s)_{s\in [0,T]}$ and continuously update and improve her actions based on these observations.

To describe the exploration step in reinforcement learning, we can randomize the actions and consider its distribution as a relaxed control. To this end, we need to assume that $(\Omega, \Fc, \{\Fc_s\}_s, \P)$ is rich enough to support a continuum of independent random variables $(Z_s)_{s \in [0, T]}$ that are uniformly distributed on $[0, 1]$ and also independent of $W$; See Theorem 1 in \cite{Sun2006} for the construction of $(Z_s)_{s \in [0, T]}$. Note that the additional randomness $(Z_s)_{s \in [0, T]}$ allows the agent to freely randomize his action at each time $s$.

Let ${\bm \pi}$ be a stochastic control that maps $(t, x, \mu) \in [0, T] \times \R^d \times \Pc_2(\R^d) \to \Pc(\Ac)$, where $\Pc(\Ac)$ is the space of probability measures on $\Ac$. At each time $s \in [0, T]$, an action $a_s: = a(s, X_s, \P_{X_s}, Z_s)$ is sampled from ${\bm \pi}(\cdot|s, X_s, \P_{X_s})$. Fix a stochastic policy ${\bm \pi}$ and an initial pair $(t, \mu) \in [0, T] \times \Pc_2(\R^d)$, let us consider the controlled McKean-Vlasov SDE
\begin{align}\label{equ:exploratory_SDE}
dX_s^{{\bm \pi}} = b(s, X_s^{{\bm \pi}}, \P_{X_s^{\bm \pi}},  a_s^{{\bm \pi}}) ds + \sigma(s, X_s^{{\bm \pi}}, \P_{X_s^{\bm \pi}}, a_s^{{\bm \pi}}) dW_s,\; X_t^{{\bm \pi}} \sim \mu, \;a_s^{\bm \pi} \sim {\bm \pi}(\cdot|s, X_s^{{\bm \pi}}, \P_{X_s^{\bm \pi}}).
\end{align}
When coefficients $b$ and $\sigma$ are Lipschitz in $(t, x, \mu) \in [0, T] \times \R^d \times \Pc_2(\R^d)$ for every action $a \in \Ac$, the existence and uniqueness of solution of \eqref{equ:exploratory_SDE} can be proved by following Theorem 5.1.1 in \cite{SC1997}. The solution to \eqref{equ:exploratory_SDE} is denoted as $X^{t, \mu, \bm \pi} = \{X_s^{t, \mu, \bm \pi}, t \leq s \leq T\}$.

To encourage the exploration in continuous time, we adopt the Shannon entropy regularizer suggested by \cite{Wangetal2021}, and the value function of the exploration formulation is given by
\begin{align}\label{equ:coupled_value_function}
J(t, \mu; {\bm \pi})& = \E\biggl[\int_t^T e^{-\beta(s-t)} \big[r(s, X_s^{{\bm \pi}}, \P_{X_s^{\bm \pi}}, a_s^{\bm \pi}) - \gamma \log {\bm \pi}(a_s^{\bm \pi}|s, X_s^{\bm \pi}, \P_{X_s^{\bm \pi}}) \big]ds\nonumber\\
&\;\;\;\;\;\; + e^{-\beta (T -t)}g( X_T^{\bm \pi}, \P_{X_T^{\bm \pi}})\Big| {X_t^{\bm \pi}} \sim \mu \biggl].
\end{align}

Due to the Shannon entropy regularization in \eqref{equ:coupled_value_function}, it is clear that the value function under the policy without a density becomes $-\infty$. Therefore, we shall restrict our attention to policies that admit densities. A stochastic policy ${\bm \pi}$ is called admissible if ${\bm \pi}$ admits a density function and is jointly measurable with respect to $(t, x, \mu, a) \in [0, T] \times \R^d \times \Pc_2(\R^d) \times \Ac$ and {\rm supp}(${\bm \pi}) = \Ac$. We denote by $\Pi$ the set of admissible (stochastic) policies ${\bm \pi}$.

We also consider the dynamics \eqref{equ:exploratory_average_SDE}, viewed intuitively as the average of the sample trajectories $X^{\bm \pi}$ in \eqref{equ:exploratory_SDE} over randomized actions.
\begin{align}\label{equ:exploratory_average_SDE}
dX_s =  b_{{\bm \pi}}(s, X_s, \P_{X_s}, {\bm \pi}(\cdot|s, X_s, \P_{X_s}))ds +  \sigma_{{\bm \pi}}(s, X_s, \P_{X_s}, {\bm \pi}(\cdot|s, X_s, \P_{X_s})) dW_s, 
\end{align}
where
\begin{align*}
&b_{{\bm \pi}}(s, x, \mu) := \int_{\Ac} b(s, x, \mu, a){\bm \pi}(a|s,x,\mu)da, \;\; \sigma_{{\bm \pi}}(s, x, \mu) := \sqrt{\int_{\Ac} \sigma\sigma\trans(s, x,\mu, a) {\bm \pi}(a|s, x, \mu) da}.
\end{align*}
The wellposedness of \eqref{equ:exploratory_average_SDE} is guaranteed under Assumption \ref{assump} (i) and its solution is denoted by $\tilde X^{\bm \pi}=\{\tilde X_s^{\bm \pi}, t \leq s \leq T\}$.
In appendix \ref{appenA}, we also rigorously prove the connection between \eqref{equ:exploratory_SDE} and \eqref{equ:exploratory_average_SDE} in the sense that ${X_s^{{\bm \pi}}}$  and $\tilde X_s^{\bm \pi}$ have the same law for every   $s \in [t, T]$ by showing that they correspond to the same martingale problem.

As $\P_{X_s^{{\bm \pi}}} = \P_{\tilde X_s^{\bm \pi}}$, the objective function in \eqref{equ:coupled_value_function} is equivalent to
\begin{align}
J(t, \mu; {\bm \pi}) & = \E\biggl[\int_t^T e^{-\beta(s-t)} \big[r_{\bm \pi}(s, \tilde X_s^{{\bm \pi}}, \P_{\tilde X_s^{\bm \pi}}) + \gamma \mathcal{E}_{\bm \pi}(s, \tilde X_s^{\bm \pi}, \P_{\tilde X_s^{\bm \pi}}))\big]ds \\
& \;\;\;\;\;\; + e^{-\beta (T -t)} g( \tilde X_T^{\bm \pi}, \P_{\tilde X_T^{\bm \pi}})\Big| \tilde X_t^{\bm \pi} \sim \mu \biggl], \nonumber
\end{align}
where we denote
\begin{align*}
r_{{\bm \pi}}(t,x,\mu):=\int_\Ac r(t, x,\mu, a){\bm \pi}(a|t,x,\mu)da, \quad \mathcal{E}_{\bm \pi}(t,x,\mu) :=-\int_\Ac \log{\bm \pi}(a|t,x,\mu) {\bm \pi}(a|t,x,\mu)
da.
\end{align*}


Then the value function associated with the admissible policy ${\bm \pi}$ satisfies the dynamic programming equation:
\begin{align}\label{equ:HJB-coupled-value-function}
&\frac{\partial J}{\partial t}(t, \mu; {\bm \pi}) - \beta J(t, \mu; {\bm \pi}) +
\E_{\xi \sim \mu}\Big[\int_{\Ac}\Big[H\big(t, \xi, \mu, a,\partial_\mu J(t, \mu; {\bm \pi})(\xi), \partial_{\color{blue}{x}}\partial_\mu J(t, \mu; {\bm \pi})(\xi)\big)\\
&\;\;\; - \gamma \log {\bm \pi}(a|t, \xi, \mu)\Big] {\bm \pi}(a|t, \xi, \mu) da \Big] =0, \nonumber
\end{align}
with the terminal condition $J(T, \mu;{\bm \pi}) = \hat g(\mu): = \int_{\R^d} g(x, \mu)\mu(dx)$.

In addition, let us denote
\begin{align*}
M_2(\mu): = \int_{\R^d} |x|^2 \mu(dx).
\end{align*}

We shall make the following regularity assumptions on coefficients and reward functions throughout the paper (see Assumption 2.1 and Remark 2.1 in \cite{FGLPS23}).

\begin{Assumption}\label{assump}
\begin{itemize}
\item[(i)] For $f\in \{b_{\bm \pi}, \sigma_{\bm \pi}\}$, the derivatives $\partial_xf(t,x,\mu)$, $\partial_{x}^2 f(t,x,\mu)$, $\partial_{\mu}f(t,x,\mu)(v)$ and $\partial_{v}\partial_{\mu} f(t,x,\mu)(v)$ exist for any $(t,x,v,\mu)\in [0,T]\times \mathbb{R}^d\times \mathbb{R}^d \times \mathcal{P}_2(\mathbb{R}^d)$, are bounded and locally Lipschitz continuous with respect to $x,\mu, v$ uniformly in $t\in[0,T]$. For all $(t,x,\mu)\in[0,T]\times\mathbb{R}^d\times\mathcal{P}_2(\mathbb{R}^d)$, we have $|f(t,x,\mu)|\leq C(1+|x|+M_2(\mu))$.

\item[(ii)] For any $t\in [0,T]$, $r_{\bm \pi }(t,\cdot)$, $\mathcal{E}_{\bm \pi}(t,\cdot)$ and $g(\cdot)\in\mathcal{C}^{2,2}(\mathbb{R}^d\times\mathcal{P}_2(\mathbb{R}^d))$.

\item[(iii)] There exists some constant $C<\infty$, such that for any $(t,x,v,\mu)\in [0,T]\times \mathbb{R}^d \times \mathbb{R}^d \times \mathcal{P}_2(\mathbb{R}^d)$, we have
\begin{align*}
&|r_{\bm\pi}(t,x,\mu)|+|\mathcal{E}_{{\bm \pi}}(t,x,\mu)|+|g(x,\mu)|\leq C(1+|x|^2+M_2(\mu)),\\
&|\partial_xr_{\bm\pi}(t,x,\mu)|+|\partial_x\mathcal{E}_{{\bm \pi}}(t,x,\mu)|+|\partial_xg(x,\mu)|\leq C(1+|x|+M_2(\mu)),\\
&|\partial_{\mu}r_{\bm\pi}(t,x,\mu)(v)|+|\partial_{\mu}\mathcal{E}_{{\bm \pi}}(t,x,\mu)(v)|+|\partial_\mu g(x,\mu)(v)|\leq C(1+|x|+|v|+M_2(\mu)),\\
&|\partial_{v}\partial_{\mu}r_{\bm\pi}(t,x,\mu)(v)|+|\partial_x^2r_{\bm\pi}(t,x,\mu)|+  |\partial_v\partial_{\mu}\mathcal{E}_{{\bm \pi}}(t,x,\mu)(v)|\\
&\quad\quad+ |\partial_x^2 \mathcal{E}_{{\bm \pi}}(t,x,\mu)|+ |\partial_v\partial_{\mu}g(x,\mu)(v)|+|\partial_x^2 g(x,\mu)|\leq C(1+M_2(\mu)).
\end{align*}
\end{itemize}
\end{Assumption}

Under Assumption \ref{assump}, Proposition $2.1$ of  \cite{FGLPS23} guarantees that the function $J(t, \mu; {\bm \pi})$ defined in \eqref{equ:coupled_value_function} is of $C^{1,2}([0,T]\times\mathcal{P}_2(\mathbb{R}^d))$.

The objective of the social planner is to maximize $J(t, \mu; {\bm \pi})$ over all admissible policies
\begin{align}\label{equ:optimal_value_function}
J^*(t, \mu) = \sup_{{\bm \pi} \in \Pi} J(t, \mu; {\bm \pi}).
\end{align}
We can derive the exploratory HJB equation for the optimal value function $J^*(t, \mu)$ by
\begin{align}\label{equ:exploratory_HJB}
&\frac{\partial J^*}{\partial t}(t, \mu) +
\int_{\R^d} \sup_{{\bm \pi} \in \Pc(\Ac)}\biggl[\int_{\Ac}\Big(H\big(t, x, \mu, a, \partial_\mu J^*(t, \mu)(x), \partial_{x}\partial_\mu J^*(t, \mu)(x)\big)\\
& \;\;\; -\;\gamma \log {\bm \pi}(a)\Big){\bm \pi}(a)da\biggl] \mu(d{x}) - \beta J^*(t, \mu) =0. \nonumber
\end{align}
The optimal policy ${\bm \pi}$ satisfies a Gibbs measure or widely-used Boltzmann policy in RL after normalization that
\begin{align}\label{equ:exploratory_optimal_policy}
{\bm \pi}^*(a|t, {x}, \mu)  = \frac{\exp\Big\{\frac{1}{\gamma}H\big(t, {x}, \mu, a, \partial_\mu J^*(t, \mu)({x}), \partial_{x}\partial_\mu J^*(t, \mu)({x})\big)\Big\}}{\int_{\Ac} \exp\Big\{\frac{1}{\gamma}H\big(t, {x}, \mu, a, \partial_\mu J^*(t, \mu)({x}), \partial_x\partial_\mu J^*(t, \mu)({x})\big)\Big\}da}.
\end{align}

Plugging \eqref{equ:exploratory_optimal_policy} into \eqref{equ:exploratory_HJB}, we get that
\begin{align}\label{equ:exploratory_HJB_2}
& \frac{\partial J^*}{\partial t}(t, \mu) + \gamma \int_{\R^d} \log \int_{\Ac} \exp\Big\{\frac{1}{\gamma}H\big(t, {x}, \mu, a, \partial_\mu J^*(t, \mu)({x}), \partial_{x}\partial_\mu J^*(t, \mu)({x})\big)\Big\} da \mu(d{x})\\
& \;\;\;-\; \beta J^*(t, \mu) = 0, \nonumber
\end{align}
with the terminal condition $J^*(T, \mu) = \hat g(\mu)$.

Recall that the goal of MFC is to find the optimal policy that maximizes $J(t, \mu; {\bm \pi})$. Let us consider the operator $\mathcal{I}: \Pi \to \Pi$ that
\begin{align}\label{equ:policy_improvemet_map}
 \mathcal{I}({\bm \pi})& = \frac{\exp\Big\{\frac{1}{\gamma}H\big(t, x, \mu, a, \partial_\mu J(t, \mu;{\bm \pi})(x), \partial_{x}\partial_\mu J(t, \mu; {\bm \pi})(x)\big)\Big\} }{\int_{\Ac}\exp\Big\{\frac{1}{\gamma}H\big(t, x, \mu, a, \partial_\mu J(t, \mu;{\bm \pi})(x), \partial_{x}\partial_\mu J(t, \mu; {\bm \pi})(x)\big)\Big\} da}.
\end{align}

\begin{Theorem}[Policy improvement result]\label{thm:policy_improvement}For any given ${\bm \pi} \in \Pi$, define ${\bm \pi}' = \mathcal{I}({\bm \pi})$, with $\mathcal{I}$ given in \eqref{equ:policy_improvemet_map}.
Then
\begin{align*}
J(t, \mu; {\bm \pi}') \geq J(t, \mu; {\bm \pi}).
\end{align*}
Moreover, if the map $\mathcal{I}$ in \eqref{equ:policy_improvemet_map} has a fixed point ${\bm \pi}^* \in \Pi$, then ${\bm \pi}^*$ is the optimal policy of \eqref{equ:optimal_value_function}.
\end{Theorem}
\textbf{Proof.} For two given admissible policies ${\bm \pi}, {\bm \pi}' \in \Pi$, and any $0 \leq t \leq T$, by applying It\^o's formula to the value function $J(s, \P_{X_s^{{\bm \pi}'}}; {\bm \pi})$ between $t$ and $T$, we get that
\begin{align}
&e^{-\beta T} J(T, \P_{X_{T}^{{\bm \pi}'}}; {\bm \pi}) - e^{-\beta t} J(t, \P_{X_{t}^{{\bm \pi}'}}; {\bm \pi}) \nonumber\\
& \;+ \E\biggl[\int_{t}^T \int_{\Ac} e^{-\beta s}\Big[r(s, X_s^{{\bm \pi}'}, \P_{X_s^{{\bm \pi}'}}, a) -\gamma \log{\bm \pi}'(a|s, X_s^{{\bm \pi}'}, \P_{X_s^{{\bm \pi}'}})\Big]{\bm \pi}'(a|s, X_s^{{\bm \pi}'}, \P_{X_s^{{\bm \pi}'}})da ds\biggl] \nonumber\\
= & \int_t^T e^{-\beta s} \Big[\frac{\partial J}{\partial t}(s, \P_{X_s^{{\bm \pi}'}}; {\bm \pi}) - \beta J(s, \P_{X_s^{{\bm \pi}'}}; {\bm \pi}) \Big]ds \nonumber\\
& \;+ \E\biggl[\int_t^T \int_{\Ac} \Big[H\big(s, X_s^{{\bm \pi}'}, \P_{X_s^{{\bm \pi}'}}, a,\partial_\mu J(s, \P_{X_s^{{\bm \pi}'}};  {\bm \pi})(X_s^{{\bm \pi}'}), \partial_{\color{blue} x}\partial_\mu J(s, \P_{X_s^{{\bm \pi}'}};  {\bm \pi})(X_s^{{\bm \pi}'})\big) \nonumber\\
&\;\;\; -   \gamma \log {\bm \pi}'(a|s, \P_{X_s^{{\bm \pi}'}}, X_s^{{\bm \pi}'})\Big]{\bm \pi}'(a|s, \P_{X_s^{{\bm \pi}'}}, X_s^{{\bm \pi}'})da ds\biggl].
\label{proof:q-function_characterization_unknown}
\end{align}
From Lemma 9 in \cite{jiazhou2022}, it follows that for any $(s, y, \mu) \in [0, T] \times \R^d \times \Pc_2(\R^d)$,
\begin{align*}
&\int_{\Ac} \Big[H(s, y, \mu, a, \partial_\mu J(s, \mu; {\bm \pi})(y), \partial_{x}\partial_\mu J(s, \mu; {\bm \pi})(y)) - \gamma \log {\bm \pi}'(a|s, y, \mu) \Big]{\bm \pi}'(a|s, y, \mu) da\\
\geq & \int_{\Ac} \Big[H(s, y, \mu, a, \partial_\mu J(s, \mu; {\bm \pi})(y), \partial_{x}\partial_\mu J(s, \mu; {\bm \pi})(y))  - \gamma \log {\bm \pi}(a|s, y, \mu)\Big]{\bm \pi}(a|s, y, \mu) da.
\end{align*}
Therefore, it holds that
\begin{align}\label{proof:policy_improvement_inequality}
&e^{-\beta T} J(T, \P_{X_{T}^{{\bm \pi}'}}; {\bm \pi}) - e^{-\beta t} J(t, \P_{X_{t}^{{\bm \pi}'}}; {\bm \pi})\\
& \;+ \E\biggl[\int_{t}^T \int_{\Ac} e^{-\beta s}\Big[r(s, X_s^{{\bm \pi}'}, \P_{X_s^{{\bm \pi}'}}, a) -\gamma \log{\bm \pi}'(a|s, X_s^{{\bm \pi}'}, \P_{X_s^{{\bm \pi}'}})\Big]{\bm \pi}'(a|s, X_s^{{\bm \pi}'}, \P_{X_s^{{\bm \pi}'}})da ds\biggl] \geq 0. \nonumber
\end{align}
Letting $\P_{X_t^{{\bm \pi}'}} = \mu$ and rearranging terms in \eqref{proof:policy_improvement_inequality} yield the desired result that
\begin{align*}
 &J(t, \mu; {\bm \pi})\leq e^{-\beta(T-t)} J(T, \P_{X_{T}^{{\bm \pi}'}}; {\bm \pi})\\
 &\;+\E\biggl[\int_{t}^T \int_{\Ac} e^{-\beta s}\Big[r(s, X_s^{{\bm \pi}'}, \P_{X_s^{{\bm \pi}'}}, a) -\gamma \log{\bm \pi}'(a|s, X_s^{{\bm \pi}'}, \P_{X_s^{{\bm \pi}'}})\Big]{\bm \pi}'(a|s, X_s^{{\bm \pi}'}, \P_{X_s^{{\bm \pi}'}})da ds\biggl]\\
 = & e^{-\beta(T-t)} \hat g(\P_{X_{T}^{{\bm \pi}'}})\\
  &\;+\E\biggl[\int_{t}^T \int_{\Ac} e^{-\beta s}\Big[r(s, X_s^{{\bm \pi}'}, \P_{X_s^{{\bm \pi}'}}, a) -\gamma \log{\bm \pi}'(a|s, X_s^{{\bm \pi}'}, \P_{X_s^{{\bm \pi}'}})\Big]{\bm \pi}'(a|s, X_s^{{\bm \pi}'}, \P_{X_s^{{\bm \pi}'}})da ds\biggl] \\
 = &J(t,\mu; {\bm \pi}').
\end{align*}

The learning procedure in Theorem \ref{thm:policy_improvement} starts with some policy ${\bm \pi}$ and produces a new policy ${\bm \pi}'$ by setting ${\bm \pi}' = \mathcal{I}({\bm \pi})$. We next estimate the distance between ${\bm \pi}$ and ${\bm \pi}'$.
\begin{Lemma} \label{lem:policy-radius-estimate}Under Assumption \ref{assump}, define ${\bm \pi}' = \mathcal{I}({\bm \pi})$ with $\mathcal{I}$ given in \eqref{equ:policy_improvemet_map} for any given ${\bm \pi} \in \Pi$, then we have
\begin{align}\label{equ:policy-radius-estimate}
0 \leq D_{KL}^{\rm average}({\bm \pi}\|{\bm \pi}'):= \int_{\R^d} D_{KL}({\bm \pi}(\cdot|t, x, \mu)\| {\bm \pi}'(\cdot|t, x, \mu)) \mu(dx) \leq \delta(\mu): =C(1 + M_2(\mu)),
\end{align}
where $C$ is a generic constant. Furthermore, if ${\bm \pi}$ is a fixed point of $\mathcal{I}$, then $D_{KL}^{\rm average}({\bm \pi}\|{\bm \pi}')=0$.
\end{Lemma}
\textbf{Proof.} By the definition of $D_{KL}$, we have that
\begin{align*}
&D_{KL}({\bm \pi}(\cdot|t, x, \mu)\| {\bm \pi}'(\cdot|t, x, \mu))= \int_{\Ac} \log \frac{{\bm \pi}(a|t, x, \mu)}{{\bm \pi}'(a|t, x, \mu)}{\bm \pi}(a|t, x, \mu)da\\
=& -\frac{1}{\gamma} \int_{\Ac} H(t, x, \mu, \partial_\mu J(t, \mu; {\bm \pi})(x), \partial_{x}\partial_\mu J(t, \mu; {\bm \pi})(x)){\bm \pi}(a|t, x, \mu) da\\
& +\log \int_{\Ac} \exp\Big\{\frac{1}{\gamma}H\big(t, x, \mu, a, \partial_\mu J(t, \mu;{\bm \pi})(x), \partial_x\partial_\mu J(t, \mu; {\bm \pi})(x)\big)\Big\}da\\
& +\int_{\Ac} \log {\bm \pi}(a|t, x, \mu) {\bm \pi}(a|t, x, \mu) da.
\end{align*}
It then follows from \eqref{equ:HJB-coupled-value-function} that
\begin{align*}
& D_{KL}^{\rm average}({\bm \pi}\|{\bm \pi}')= \int_{\R^d} D_{KL}({\bm \pi}(\cdot|t, x, \mu)\| {\bm \pi}'(\cdot|t, x, \mu))\mu(dx)\\
=& \frac{\partial J}{\partial t}(t, \mu; {\bm \pi}) - \beta J(t, \mu; {\bm \pi})\\
&\; + \int_{\R^d}\log \int_{\Ac} \exp\Big\{\frac{1}{\gamma}H\big(t, x, \mu, a, \partial_\mu J(t, \mu;{\bm \pi})(x), \partial_{x}\partial_\mu J(t, \mu; {\bm \pi})(x)\big)\Big\}da\mu(dx).
\end{align*}
We then estimate $D_{KL}^{\rm average}({\bm \pi}\|{\bm \pi}')$. Under Assumption \ref{assump}, the proof of Proposition 2.1 in \cite{FGLPS23} implies that
\begin{align*}
\sup_{t \in [0, T]}\Big\{|\frac{\partial J}{\partial t}(t, \mu; {\bm \pi})| + |J(t, \mu; {\bm \pi})| + |\partial_\mu J(t, \mu; {\bm \pi})({x})| + |\partial_{x}\partial_\mu J(t, \mu; {\bm \pi})({x})| \Big\} & \leq C(1 + M_2(\mu)).
\end{align*}
Consequently, the desired result holds that
$D_{KL}^{\rm average}({\bm \pi}\|{\bm \pi}') \leq C(1 + M_2(\mu))$.

\section{q-Functions for Continuous Time Mean-field Control}\label{sec:q-function}
The aim of this section is to examine the continuous time analogue of the discrete time IQ-function and Q-learning for mean-field control problems (see \cite{GGWX23, CLT23}). In particular, in the framework of continuous time McKean-Vlasov control, it is an interesting open problem that what is the correct definition of the integrated q-function comparing with the q-function for a single agent's control problem in \cite{jiazhou2022} and the IQ-function in the discrete time framework in \cite{GGWX23, CLT23}? In addition, it is important to explore how can one utilize the learnt integrated q-function to learn the optimal policy?

\subsection{Soft Q-learning for Mean-field Control}\label{sec:Q}
To better elaborate our definitions of continuous time q-functions for mean-field control problems, let us first detour in this subsection to discuss the correct definition and results of Q-learning with entropy regularizer (soft Q-learning) in the discrete time framework.

We consider a mean-field Markov Decision Process (MDP) $X = \{X_t, t =0, 1, \ldots, T\}$ with a finite state space $\Xc$ and a finite action space $\Ac$, and a transition probability of mean-field type $\P(X_{t +1} = x' | X_{t} = x, \P_{X_t} = \mu, a_t = a) =: p(x'|x, \mu, a)$. At each time step $t$, an action is sampled from a stochastic policy ${\bm \pi}$. The social planner's expected total reward is $\E[\sum_{t=0}^{T-1} \beta^t r(X_t, \P_{X_t}, a_t) + g(X_T, \P_{X_T})]$, with $\beta: = \exp(-\alpha) \in (0, 1)$.

The value function at each time $t$ associated with a given policy ${\bm \pi}$ is defined as
\begin{align*}
J(t, \mu; \{{\bm \pi}_s\}_{s \geq t}) = \E\Big[\sum_{s=t}^{T - 1} \beta^{(s - t)} [r(X_s^{\bm \pi}, \P_{X_s^{\bm \pi}}, a_s^{\bm \pi}) - \gamma \log {\bm \pi}(a_s^{\bm \pi}|s, X_s^{\bm \pi}, \P_{X_s^{\bm \pi}})] + \beta^{(T-t)}g(X_T^{\bm \pi}, \P_{X_T^{\bm \pi}})\Big],
\end{align*}
with $\P_{X_t^{\bm \pi}} = \mu$. The Bellman equation for $J$ is given by
\begin{align}
J(t, \mu; \{{\bm \pi}_s\}_{s \geq t}) = \E\big[r(X_t^{\bm \pi}, \mu, a_t^{\bm \pi}) - \gamma \log {\bm \pi}(a_t^{\bm \pi}|s, X_t^{\bm \pi}, \mu)\big] + \beta J(t + 1, \P_{X_{t + 1}^{\bm \pi}}; \{{\bm \pi}_s\}_{s \geq t + 1}).
\end{align}
When the social planner takes the policy ${\bm h}$ at time $t$, and then the policy $\{\bm \pi_s\}_{s \geq t + 1}$ afterwards, the integrated Q-function (see \cite{GGWX23} without the entropy regularizer) is defined by
\begin{align}
& Q(t, \mu, {\bm h}; \{{\bm \pi}_s\}_{s \geq t + 1}) \nonumber\\
= &  \E\big[r(X_t^{\bm h},  \P_{X_t^{\bm h}}, a_t^{\bm h}) - \gamma \log {\bm h}(a_t^{\bm h}|t, X_t^{\bm h}, \P_{X_t^{\bm h}})\big] \nonumber\\
& + \beta \E\biggl[\sum_{s=t+1}^{T-1} \beta^{s-(t+1)} \big[r(X_s^{{\bm \pi}}, \P_{X_s^{{\bm \pi}}}, a_s^{{\bm \pi}}) - \gamma\log {\bm \pi}(a_s^{\bar{\bm \pi}}|s, X_s^{{\bm \pi}}, \P_{X_s^{{\bm \pi}}})\big]\Big|X_{t+1}^{{\bm \pi}} \sim \P_{X_{t+1}^{\bm h}}\biggl] \nonumber\\
= & \E\big[r(X_t^{\bm h},  \P_{X_t^{\bm h}}, a_t^{\bm h}) - \gamma \log {\bm h}(a_t^{\bm h}|t, X_t^{\bm h}, \P_{X_t^{\bm h}})\big] + \beta J(t+1,  \P_{X_{t+1}^{\bm h}}; \{{\bm \pi}_s\}_{s \geq t + 1}), \label{appendix:definition-Q}
\end{align}
with the terminal condition $Q(T, \mu, {\bm h}) = \E_{\xi \sim \mu}[g(\xi, \mu)]$.

Next, we consider the optimal value function $J^*(t, \mu): = \sup_{\{{\bm \pi}_s\}_{s \geq t + 1}}J(t, \mu; \{{\bm \pi}_s\}_{s \geq t + 1})$ and the optimal Q-function and $Q^*(t, \mu, {\bm h}) = \sup_{\{{\bm \pi}_s\}_{s \geq t + 1}}Q(t, \mu; {\bm h}; \{{\bm \pi}_s\}_{s \geq t + 1})$ associated with the optimal policy ${\bm \pi}^*$. First, the DPP or the Bellman equation for $J^*$ is
\begin{align}\label{dppJ}
J^*(t, \mu) = \sup_{{\bm \pi}} \Big\{\E_{\mu, {\bm \pi}}\big[r(\xi, \mu, a_t^{\bm \pi}) - \gamma \log {\bm \pi}(a_t^{\bm \pi}|t, \xi, \mu)\big] + \beta J^*(t + 1, \P_{X_{t + 1}^{t, \mu, \bm \pi}})\Big\},
\end{align}
where for any measurable function $f$, $\mu \in \Pc_2(\R^d)$ and ${\bm h} \in \Pi$, we denote
\begin{align}\label{Emuh}
\E_{\mu, {\bm h}}[f(\xi, a^{\bm h})]: = \E_{\xi \sim \mu, a^{\bm h} \sim {\bm h}(\cdot|t, \xi, \mu)}[f(\xi, a^{\bm h})] = \int_{\R^d}\int_{\Ac} f(x, a ){\bm h}(a|t, x, \mu)da\mu(dx).
\end{align}

From \eqref{appendix:definition-Q} and \eqref{dppJ}, we have the relation between the optimal value function and the optimal Q-function
\begin{align} \label{appendix:relation-J-Q}
J^*(t, \mu) = \sup_{{\bm h}} Q^*(t, \mu, {\bm h}).
\end{align}
Substituting \eqref{appendix:relation-J-Q} to the Bellman equation for $J^*$, we obtain that
\begin{align}
J^*(t, \mu) &= \sup_{{\bm \pi}} \Big\{\E_{\mu, {\bm \pi}}\big[r(\xi, \mu, a_t^{\bm \pi}) - \gamma \log {\bm \pi}(a_t^{\bm \pi}|t, \xi, \mu)\big] + \beta \sup_{{\bm h}'} Q^*(t + 1, \P_{X_{t + 1}^{t, \mu, \bm \pi}}; {\bm h}')\Big\} \nonumber\\
& = \sup_{{\bm \pi}} \Big\{\sum_{x \in \Xc}\sum_{a \in \Ac}\big(r(x, \mu, a) - \gamma \log {\bm \pi}(a|t, x, \mu)\big) {\bm \pi}(a|t, x, \mu) \mu(x) \nonumber\\
& + \beta \sup_{{\bm h}'} Q^*(t + 1, \Phi(t, \mu, {\bm \pi}), {\bm h}')\Big\}, \label{equ:Bellman-optimalJ}
\end{align}
where $\Phi$, which characterizes the evolution of $\P_{X_t^{\bm \pi}}$ over time, is defined by $\Phi(t, \mu, {\bm h})(x') = \sum_{x \in \Xc} \sum_{a \in \Ac} p(x'|x, \mu, a){\bm h}(a|t, x, \mu)\mu(x)$.

To derive the optimal policy, we need to find the candidate policy that achieves $\sup_{{\bm h}}  Q^*(t, \mu, {\bm h})$ for any $\mu \in \Pc_2(\Xc)$ and $t \in {0, 1, \ldots, T -1}$, or equivalently we can search for the optimal policy by solving the optimization problem on the right hand side of \eqref{equ:Bellman-optimalJ}. However, due to the nonlinear dependence of $\sup_{{\bm h}'}Q^*(t + 1, \Phi(t, \P_{X_t^{\bm \pi}}, {\bm \pi}), {\bm h}')$ in ${\bm \pi}$, it is impossible to obtain an explicit expression of ${\bm \pi}^*$. Therefore, for McKean-Vlasov control problems in the mean-field model, we reveal that the introduction of the entropy regularizer does not provide any help to derive the distribution of the optimal policy ${\bm \pi}^*$, which differs significantly from the soft Q-learning for the single agent's stochastic control problem, see \cite{jiazhou2022}.

Moreover, taking the supremum over all $\{{\bm \pi}_s\}_{s \geq t + 1}$ in \eqref{appendix:definition-Q}, and substituting \eqref{appendix:relation-J-Q} to \eqref{appendix:definition-Q}, we can heuristically obtain the DPP or the Bellman equation for the optimal IQ-function that
\begin{align} \label{appendix:Bellaman-optimal-Q}
Q^*(t, \mu, {\bm h}) = \E_{\mu, {\bm h}}\big[r(\xi,  \mu, a_t^{\bm h}) - \gamma \log {\bm h}(a_t^{\bm h}|t, \xi, \mu)\big] + \beta \sup_{{\bm h}'} Q^*(t + 1, \P_{X_{t + 1}^{t, \mu, \bm h}}, {\bm h}'),
\end{align}
with the terminal condition $Q^*(T, \mu, {\bm h}) = \E_{\xi \sim \mu}[g(\xi, \mu)]$.

Note that when $\gamma = 0$, \eqref{appendix:Bellaman-optimal-Q} coincides with the Bellman equation for the optimal IQ-function established in \cite{GGWX23}, in which the correct definition of the IQ-function $Q(t, \mu, {\bm h})$ defined on the distribution of both the population and the action is proposed for MFC such that the time consistency and DPP hold. In \eqref{appendix:Bellaman-optimal-Q} for the learning MFC problem with the entropy regularizer, it is straightforward to modify arguments in  \cite{GGWX23} and similarly prove the correct DPP for the optimal IQ-function with an additional entropy term associated with ${\bm h}$.

On the other hand, when there is no mean-field interaction, i.e., there is no population distribution in the transition probability $p$ or in the reward $r$, the terminal payoff $g$ and the stochastic policy ${\bm \pi}$, the problem degenerates to the single-agent Q learning. We can use either the single-agent Q-function denoted by $Q^*_{\rm single}$  or the IQ-function to learn the optimal policy. In fact, these two Q-functions are related by the equation
\begin{align}\label{appendix:relation-Q-Qsingle}
Q^*(t, \mu, {\bm h}) = \sum_{x \in \Xc} \sum_{a \in \Ac} \big(Q_{\rm single}^* (t, x, a)  - \gamma \log {\bm h}(a|t, x)\big){\bm h}(a|t, x)\mu(x),
\end{align}
which implies that the optimal policy that attains $\sup_{{\bm h}} Q^*(t, \mu, {\bm h})$ can be explicitly written by
\begin{align}\label{appendix:single-agent-optimal-policy}
{\bm \pi}^*(a|t, x) = \frac{\exp\{\frac{1}{\gamma} Q^*_{\rm single} (t, x, a)\}}{\sum_{a \in \Ac}\exp\{\frac{1}{\gamma} Q^*_{\rm single} (t, x, a)\}},
\end{align}
which has been shown in \cite{jiazhou2022}.

According to \eqref{appendix:relation-Q-Qsingle}, we see that \eqref{appendix:Bellaman-optimal-Q} can be reformulated in terms of $Q_{\rm single}^*$ that
\begin{align}
Q^*_{\rm single}(t, x, a) &= r(t, x , a) + \beta \sum_{x' \in \Xc} \gamma \log \sum_{a \in \Ac} \exp\big\{\frac{1}{\gamma} Q^*_{\rm single} (t + 1, x', a)\big\} p(x'|t, x, a) \nonumber\\
& =  r(t, x , a) + \beta \E\big[\gamma \log \sum_{a \in \Ac} \exp\big\{\frac{1}{\gamma} Q^*_{\rm single} (t + 1,X_{t + 1}^{t, x, a}, a)\big\}\big| X_t = x, a_t = a\big],
\end{align}
and that \eqref{appendix:relation-J-Q} can be expressed in terms of $Q_{\rm single}^*$ and $J^*_{\rm single}$ that
\begin{align}\label{appendix:single-agent-relation-J-Q}
J^*_{{\rm single}}(t, x) = \sup_{{\bm \pi}} \big\{\sum_{a \in \Ac} \big(Q_{\rm single}^* (t, x, a)  - \gamma \log {\bm \pi}(a|t, x)\big){\bm \pi}(a|t, x)\big\}.
\end{align}
Equations \eqref{appendix:single-agent-optimal-policy}-\eqref{appendix:single-agent-relation-J-Q} are consistent with results for the single-agent soft Q-learning, see e.g. \cite{SuttonBarto2018} and \cite{jiazhou2022}.

\subsection{Two Continuous Time q-functions}
Due to the large population of interacting agents as well as the technical issue discussed in Remark \ref{rmk:reason-h} below, we need to consider the randomized action ${\bm h} \in \Pi$ on the interval $[t, t+ \Delta t)$ in the IQ-function in the McKean-Vlasov control framework (see \cite{GGWX23}), which differs significantly from the perturbed policy with a constant action in \cite{jiazhou2022} for a single agent. Therefore, let us consider a ``perturbed policy" $\bar {\bm \pi} \in \Pi$, which takes ${\bm h} \in \Pi $ on $[t, t+ \Delta t)$, and then ${\bm \pi} \in \Pi$ on $[t+\Delta t, T)$.  By Lemma \ref{lem:policy-radius-estimate}, it is sufficient to restrict ${\bm h}$ within the ball $B_{\delta(\mu)}({\bm \pi})= \{{\bm h} \in \Pi: D_{KL}^{average}({\bm \pi}||{\bm h}) \leq \delta(\mu)\}$ centered at ${\bm \pi}$ that significantly  reduces the searching space of policies ${\bm h}$.

Denote $\rho_s^{t, \mu, {\bm \pi}}: = \P_{X_s^{t, \mu, {\bm \pi}}}$ for any ${\bm \pi} \in \Pi$ for notational simplicity. The state process $X_s^{t, \mu, \bar{\bm \pi}}$ on $[t, T)$ is governed by
\begin{align*}
 dX_s^{t, \mu, \bar {\bm \pi}} &= b(s, X_s^{t, \mu, \bar {\bm \pi}},  \rho_s^{t, \mu, \bar {\bm \pi}}, {a}_s^{\bm h})ds + \sigma(s, X_s^{t, \mu, \bar {\bm \pi}},  \rho_s^{t, \mu, \bar {\bm \pi}}, {a}_s^{\bm h}) dW_s, \; s \in [t, t+\Delta t), \; {X_{t}^{t, \mu, \bar {\bm \pi}}} \sim \mu,\\
 dX_s^{t, \mu, \bar {\bm \pi}} &= b(s, X_s^{t, \mu, \bar {\bm \pi}},  \rho_s^{t, \mu, \bar {\bm \pi}}, {a}_s^{\bm \pi})ds + \sigma(s, X_s^{t, \mu, \bar {\bm \pi}},  \rho_s^{t, \mu, \bar {\bm \pi}}, {a}_s^{\bm \pi}) dW_s, \; s \in [t+\Delta t, T), \; {X_{t+ \Delta t}^{t, \mu, \bar {\bm \pi}}} \sim \rho_{t + \Delta t}^{t, \mu, {\bm h}}.
\end{align*}

Based on the definition in \eqref{appendix:definition-Q}, we can first consider the discrete time integrated $Q$-function defined on $[0, T] \times \Pc_2(\R^d) \times {B_{\delta(\mu)}({\bm \pi})} \times \Pi$ with the time interval $\Delta t$ and the entropy regularizer that
\begin{align}  \label{equ:def_Q}
& Q_{\Delta t} (t, \mu, {\bm h}; {\bm \pi}) \\
=& \E\biggl[\int_t^{t + \Delta t} \int_{\Ac}e^{-\beta(s-t)}\Big[r(s, X_s^{t, \mu, \bar{\bm \pi}}, \rho_s^{t, \mu, \bar{\bm \pi}}, a) - \gamma \log {\bm h}(a|s, X_s^{t, \mu, \bar{\bm \pi}}, \rho_s^{t, \mu, \bar{\bm \pi}})\Big] {\bm h}(a|s,  X_s^{t, \mu, \bar{\bm \pi}}, \rho_s^{t, \mu, \bar{\bm \pi}}) da ds\nonumber\\
& \;+ \int_{t + \Delta t}^T \int_{\Ac} e^{-\beta(s-t)} \Big[r(s,  X_s^{t, \mu, \bar{\bm \pi}}, \rho_s^{t, \mu, \bar{\bm \pi}}, a) - \gamma \log {\bm \pi}(a|s,  X_s^{t, \mu, \bar{\bm \pi}}, \rho_s^{t, \mu, \bar{\bm \pi}})\Big] {\bm \pi}(a|s,  X_s^{t, \mu, \bar{\bm \pi}}, \rho_s^{t, \mu, \bar{\bm \pi}})da ds \nonumber\\
& \;+ e^{-\beta(T-t)} g(X_T^{t, \mu, \bar{\bm \pi}}, \rho_T^{t, \mu, \bar{\bm \pi}}) \Big| {X_t^{t, \mu, \bar{\bm \pi}}} \sim \mu\biggl] \nonumber\\
 =& \E\biggl[\int_t^{t + \Delta t} \int_{\Ac}e^{-\beta(s-t)}\Big[r(s, X_s^{t, \mu, \bar{\bm \pi}}, \rho_s^{t, \mu, \bar{\bm \pi}}, a) - \gamma \log {\bm h}(a|s, X_s^{t, \mu, \bar{\bm \pi}}, \rho_s^{t, \mu, \bar{\bm \pi}})\Big] {\bm h}(a|s,  X_s^{t, \mu, \bar{\bm \pi}}, \rho_s^{t, \mu, \bar{\bm \pi}}) da\biggl] \nonumber\\
 & \;+ e^{-\beta \Delta t} J(t + \Delta t, \rho_{t+\Delta t}^{t, \mu, \bar{\bm \pi}}; {\bm \pi}) - J(t, \mu; {\bm \pi}) + J(t, \mu; {\bm \pi}) \nonumber\\
 =& 
 \E\biggl[\int_t^{t + \Delta t} \int_{\Ac}e^{-\beta(s-t)}\Big[H\big(s, X_s^{t, \mu, {\bm h}}, \rho_s^{t, \mu, {\bm h}}, a, \partial_\mu J(s, \rho_s^{t, \mu, {\bm h}}; {\bm \pi})(X_s^{t, \mu, {\bm h}}), \partial_x\partial_\mu J(s,  \rho_s^{t, \mu, {\bm h}}; {\bm \pi})(X_s^{t, \mu, {\bm h}})\big) \nonumber\\
 & - \gamma \log {\bm h}(a|s, X_s^{t, \mu, \bar{\bm \pi}}, \rho_s^{t, \mu, \bar{\bm \pi}})\Big] {\bm h}(a|s, X_s^{t, \mu, \bar{\bm \pi}}, \rho_s^{t, \mu, \bar{\bm \pi}}) da ds\biggl] \nonumber\\
 & + \int_t^{t + \Delta t} \Big[\frac{\partial J}{\partial t} (s, \rho_s^{t, \mu, {\bm h}}; {\bm \pi}) - \beta J(s, \rho_s^{t, \mu, {\bm h}}; {\bm \pi}) \Big]ds +  J(t, \mu; {\bm \pi}) \nonumber\\
=& J(t, \mu; {\bm \pi}) + \Delta t \biggl[ \frac{\partial J}{\partial t}(t, \mu; {\bm \pi}) - \beta J(t, \mu; {\bm \pi})  + \E_{\xi \sim \mu}\Big[\int_{\Ac}\Big( H\big(t, \xi, \mu, a, \partial_\mu J(t, \mu; {\bm \pi})(\xi), \partial_x\partial_\mu J(t, \mu; {\bm \pi})(\xi)\big) \nonumber\\
& \; - \gamma \log {\bm h}(a|t, \xi, \mu)\Big){\bm h}(a|t, \xi, \mu)da\Big]\biggl] + o(\Delta t), \nonumber
\end{align}
where we have applied It\^o's formula to $e^{-\beta \Delta t} J(t + \Delta t, \rho_{t+\Delta t}^{t, \mu, \bar{\bm \pi}}; {\bm \pi}) - J(t, \mu; {\bm \pi})$ in the third equality.

\begin{Definition}\label{def:coupled-q-function}
Fix a stochastic policy ${\bm \pi} \in \Pi$. For any $(t, \mu, {\bm h}) \in [0, T] \times \Pc_2(\R^d) \times B_{\delta(\mu)}({\bm \pi})$, we define the continuous time integrated q-function by
\begin{align*}
q(t, \mu, {\bm h}; {\bm \pi}) & := \frac{\partial J}{\partial t}(t, \mu; {\bm \pi}) - \beta J(t, \mu; {\bm \pi})+ \E_{\mu, {\bm h}}\Big[H\big(t, \xi, \mu, a^{\bm h}, \partial_\mu J(t, \mu; {\bm \pi})(\xi), \partial_x\partial_\mu J(t, \mu; {\bm \pi})(\xi)\big)\Big]\\
& +\gamma \E_{\xi \sim \mu}[\mathcal{E}_{\bm h}(t, \xi, \mu)].
\end{align*}
\end{Definition}
We also call the integrated q-function associated with the optimal policy ${\bm \pi}^*$ in \eqref{equ:exploratory_optimal_policy} as the optimal integrated q-function, which is defined by
\begin{align*}
q^*(t, \mu, {\bm h}) & := \frac{\partial J^*}{\partial t}(t, \mu) - \beta J^*(t, \mu)+\E_{\mu, {\bm h}}\Big[H\big(t, \xi, \mu, a^{\bm h}, \partial_\mu J^*(t, \mu)(\xi), \partial_x\partial_\mu J^*(t, \mu)(\xi)\big)\Big]\\
& +\gamma \E_{\xi \sim \mu}[\mathcal{E}_{\bm h}(t, \xi, \mu)].
\end{align*}

\begin{Remark}
In the framework of learning MFC, with a fixed small time step size $\Delta t$, the integrated Q-function $Q_{\Delta t}(t, \mu, {\bm h}; {\bm \pi})$  is related to our integrated q-function in the following sense:
\begin{align*}
Q_{\Delta t}(t, \mu, {\bm h}; {\bm \pi}) \approx J(t, \mu; {\bm \pi}) + q(t, \mu, {\bm h}; {\bm \pi}) \Delta t.
\end{align*}
That is, $q(t, \mu, {\bm h}; {\bm \pi})$ is the first-order derivative of the integrated Q-function $Q_{\Delta t}(t, \mu, {\bm h}; {\bm \pi})$ with respect to $\Delta t$.

Contrary to the discrete time Q-learning algorithm that learns $Q_{\Delta t}$, the q-learning algorithm learns zeroth-order and first-order terms of $Q_{\Delta t}$ simultaneously. These two terms are independent of $\Delta t$ and therefore robust to the choice of the time discretization in implementations (see the discussion and numerical comparison results in \cite{jiazhou2022} for the single agent's control problem).
\end{Remark}

\begin{Definition}\label{defqe}
If there exists a function $q_e: [0, T] \times \R^d \times \Pc_2(\R^d) \times \Ac \to \R$ such that
\begin{align}\label{relationq_dq_cbarq}
q(t, \mu, {\bm h}; {\bm \pi}) - {\gamma \E_{\xi \sim \mu}[\mathcal{E}_{\bm h}(t, \xi, \mu)]}& =  \E_{\mu, {\bm h}} [q_e(t, \xi, \mu, a^{\bm h}; {\bm \pi})],
\end{align}
it is called the {\it essential} q-function, which plays the essential role in the policy improvement and the characterization of the optimal policy.

Contrary to the integrated $q$-function, the function $q_e$ above is defined on $[0, T] \times \R^d \times \Pc_2(\R^d) \times \Ac$, independent of ${\bm h}$. The following result ensures the existence of an essential q-function in our current framework.

\begin{Lemma}
There exists at least one essential q-function $q_e$ given in Definition \ref{defqe}.
\end{Lemma}
\textbf{Proof.}
Note from Definition \ref{defqe} that
\begin{align*}
& q(t, \mu, {\bm h}; {\bm \pi}) - {\gamma \E_{\xi \sim \mu}[\mathcal{E}_{\bm h}(t, \xi, \mu)]}\\
= & \frac{\partial J}{\partial t}(t, \mu; {\bm \pi}) - \beta J(t, \mu; {\bm \pi})+ \E_{\mu, {\bm h}}\Big[H\big(t, \xi, \mu, a^{\bm h}, \partial_\mu J(t, \mu; {\bm \pi})(\xi), \partial_x\partial_\mu J(t, \mu; {\bm \pi})(\xi)\big)\Big],
\end{align*}
is linear in ${\bm h}$.
Therefore, we can always define
\begin{align}\label{form-essential-q}
q_e(t, x, \mu, a; {\bm \pi}) :=& \frac{\partial J}{\partial t}(t, \mu; {\bm \pi}) - \beta J(t, \mu; {\bm \pi})\\
& + H\big(t, x, \mu, a, \partial_\mu J(t, \mu; {\bm \pi})(x), \partial_x\partial_\mu J(t, \mu; {\bm \pi})(x)\big). \nonumber
\end{align}

Then $\mathcal{I}$ defined in \eqref{equ:policy_improvemet_map} can be written in terms of the essential q-function $q_e$ that
\begin{align*}
\mathcal{I}({\bm \pi}) = \frac{\exp({\frac{1}{\gamma}}q_e(t, x, \mu, a; {\bm \pi}))}{\int_{\Ac}\exp({\frac{1}{\gamma}}q_e(t, x, \mu, a; {\bm \pi}))da}.
\end{align*}
\end{Definition}

\begin{Remark}\label{continuousRM}

It is worth noting that the integrated Q-function $Q_{\Delta t}(t, \mu, {\bm h}; {\bm \pi}) $ in general discrete time MFC problems has an implicit and nonlinear dependence on ${\bm h}$, which has three direct consequences: (i) in contrast to single-agent control problems where the optimal policy is in the form of Gibbs measure \eqref{appendix:single-agent-optimal-policy}, the optimal policy ${\bm \pi}^*$ in the discrete time MFC Q-learning cannot be expressed in terms of Gibbs measure and does not exhibit any explicit distribution;  (ii) the relationship in \eqref{appendix:relation-Q-Qsingle} does not hold in general even when there is no entropy that $\gamma=0$; (iii) the counterpart of the essential q-function in the discrete time framework does not exist. Indeed, one cannot define a discrete time Q-function only depending on $a$ (independent of $\bm h$) that satisfies the DPP, see the discussions in \cite{GGWX23}.

By contrast, as the first order derivative of $Q_{\Delta t}(t, \mu, {\bm h}; {\bm \pi})$  with respect to time $\Delta t$, the integrated q-function $q(t, \mu, {\bm h}; {\bm \pi})$ modified by the entropy term ${\gamma \E_{\xi \sim \mu}[\mathcal{E}_{\bm h}(t, \xi, \mu)]}$ has the nice linear dependence on the control policy ${\bm h}$ in view of the integral relationship \eqref{relationq_dq_cbarq}. As a consequence, solving the optimization problem $\sup_{{\bm h}}\{J(t, \mu; {\bm \pi}^*) + q(t, \mu, {\bm h}; {\bm \pi}^*) \Delta t\}$ leads to an explicit form of ${\bm \pi}^*$ in \eqref{equ:exploratory_optimal_policy}. This illustrates another advantage of working in continuous time framework for MFC as it is more convenient to characterize the optimal policy ${\bm \pi}^*$ and establish the relationship between the optimal policy ${\bm \pi}^*$ and the q-function to devise algorithms for policy improvement. We also note that the integrated q-function $q(t, \mu, {\bm h}; {\bm \pi})$ is more natural for the name ``integrated function" as it (modified by the entropy) can be expressed as a linear double-integral in  \eqref{relationq_dq_cbarq} that explicitly integrates the distribution of the state and the action of the population, while the discrete time integrated Q-function  $Q_{\Delta t}(t, \mu, {\bm h}; {\bm \pi}) $ aggregates the distribution in an implicit manner.

Finally, we stress that the integrated q-function is unique, but the essential q-function is not. In fact, any function $\kappa(x, \mu)$ satisfying $\int_{\R^d} \kappa(x, \mu) = 0$ can be added to \eqref{form-essential-q} and still be an essential q-function.
\end{Remark}

\section{Weak Martingale Characterizations}\label{sec:martingale_characterization}

The first result below gives a characterization of the integrated q-function associated with a given policy ${\bm \pi} \in \Pi$ under the assumption that the value function $J$ is known.

\begin{Theorem}[Characterization of the integrated q-function] \label{thm:known-q-function} Let ${\bm \pi} \in \Pi$, its value function $J$ and a continuous function $\hat q: [0, T] \times \Pc_2(\R^d) \times B_{\delta(\mu)}({\bm \pi})  \to \R$ be given.
 Then $\hat q$ is the integrated q-function if and only if $\hat q$ satisfies
\begin{align}\label{equ:qc-function_constraint}
\hat q(t, \mu, {\bm \pi}) =0,
\end{align}
and for any $(t, \mu, {\bm h}) \in [0, T] \times \Pc_2(\R^d) \times B_{\delta(\mu)}({\bm \pi})$, the value of
\begin{align*}
&e^{-\beta s}  J(s, \P_{X_s^{t, \mu, \bm h}}; {\bm \pi})- e^{-\beta t}  J(t, \mu; {\bm \pi})+ \int_{t}^s e^{-\beta t'}\Big[\hat r(t', \P_{X_{t'}^{\bm h}}, {\bm h}) -\hat q (t', \P_{X_{t'}^{\bm h}}, {\bm h}) {+}{\gamma \E[\mathcal{E}_{\bm h}(t, X_{t'}^{\bm h}, \P_{X_{t'}^{\bm h}})]}\Big]dt'
\end{align*}
is $0$, where $\{X_s^{\bm h}, t \leq s \leq T\}$ is the solution to \eqref{equ:exploratory_SDE} under the policy ${\bm h}$ with $\P_{X_t^{\bm h}} = \mu$, and $\hat r: [0, T] \times \Pc_2(\R^d) \times B_{\delta(\mu)}({\bm \pi}) \to \R$ is the aggregated reward defined by
$\hat r(t, \mu, {\bm h}) = \E_{\mu, {\bm h}}\big[r(t, \xi, \mu, a^{\bm h})\big]$.
\end{Theorem}
\textbf{Proof.} By \eqref{equ:HJB-coupled-value-function} and Definition \ref{def:coupled-q-function}, it is a direct consequence that \eqref{equ:qc-function_constraint} holds.

Applying It\^o's formula to $J(t', \P_{X_{t'}^{\bm h}}; {\bm \pi})$ between $t$ and $s$, $0 \leq t < s \leq T$, we get that
\begin{align*}
& e^{-\beta s}J(s, \P_{X_s^{\bm h}}; {\bm \pi}) - e^{-\beta t}  J(t, \mu; {\bm \pi})+ \int_{t}^s e^{-\beta t'} \Big[\hat r(t', \P_{X_{t'}^{\bm h}}, {\bm h}) -\hat q(t', \P_{X_{t'}^{\bm h}}, {\bm h}) + {\gamma \E[\mathcal{E}_{\bm h}(t, X_{t'}^{\bm h}, \P_{X_{t'}^{\bm h}})]}\Big]  dt'\\
=& \int_t^s e^{-\beta t'} \biggl\{\frac{\partial}{\partial t}J(t', \P_{X_{t'}^{\bm h}}; {\bm \pi}) - \beta J(t', \P_{X_{t'}^{\bm h}}; {\bm \pi})\\
&\;+ \int_{\
R^d}\int_{\Ac} H\big(t', X_{t'}^{\bm h}, \P_{X_{t'}^{\bm h}}, a, \partial_\mu J(t', \P_{X_{t'}^{\bm h}}; {\bm \pi})( X_{t'}^{\bm h}), \partial_{\color{blue} x}\partial_\mu J(t', \P_{X_{t'}^{\bm h}}; {\bm \pi})(X_{t'}^{\bm h})\big){\bm h}(a|t', X_{t'}^{\bm h}, \P_{X_{t'}^{\bm h}})da \P_{X_{t'}^{\bm h}}(dx)\\
&\;{+} {\gamma \E[\mathcal{E}_{\bm h}(t, X_{t'}^{\bm h}, \P_{X_{t'}^{\bm h}})]} -\hat q(t',  \P_{X_{t'}^{\bm h}}, {\bm h})\biggl\} dt'\\
=& \int_t^s  e^{-\beta t'} \Big[q(t', \P_{X_{t'}^{\bm h}}, {\bm h}; {\bm \pi}) - \hat q(t', \P_{X_{t'}^{\bm h}}, {\bm h})\Big]dt'
= 0.
\end{align*}

Conversely, we need to show
\begin{align}\label{proof:weak_martingale1}
\int_t^s  e^{-\beta t'} \Big[q(t', \P_{X_{t'}^{\bm h}}, {\bm h}; {\bm \pi}) - \hat q(t', \P_{X_{t'}^{\bm h}}, {\bm h})\Big]dt'=0
\end{align}
implies that $q(t, \mu, {\bm h}; {\bm \pi})\equiv \hat q(t, \mu, {\bm h})$, which will be proved by contradiction. Denote $f(t, \mu, {\bm h}) = q(t, \mu, {\bm h}; {\bm \pi}) - \hat q(t, \mu, {\bm h})$.  Then $f$ is a continuous function that maps $[0, T]  \times \Pc_2(\R^d) \times B_{\delta(\mu)}({\bm \pi})$ to $\R$. Suppose that the claim does not hold. Then there exists $(t^*, \mu^*, {\bm h}^*) \in [0, T] \times \Pc_2(\R^d) \times B_{\delta(\mu)}({\bm \pi})$ and $\epsilon > 0$ such that $f(t^*, \mu^*, {\bm h}^*) > \epsilon$. As $f$ is continuous, there exists $\delta >0$ such that when $\max\{|t-t^*|, \Wc_2(\mu, \mu^*)\} < \delta$, $f(t, \mu, {\bm h}^*) > \frac{\epsilon}{2}$. Now let us consider the process $X_s^{{\bm h}^*}$ starting from $(t^*, \mu^*)$, i.e., $X_s^{{\bm h}^*}, t^* \leq s \leq T\}$ with $X_{t^*}^{\bm {h}^*}\sim \mu^*$. Define
\begin{align*}
\tau := \inf\{t > t^*: \Wc_2(\P_{X_t^{{\bm h}^*}}, \mu^*) > \delta\} \wedge (t^* + \delta).
\end{align*}
Then it holds that  $\int_{t^*}^\tau  e^{-\beta t'}f(t', \P_{X_{t'}^{{\bm h}^*}}, {\bm h}^*) dt'  > 0$, which contradicts with \eqref{proof:weak_martingale1}, and our conclusion holds.


\begin{Remark} \label{rmk:reason-h} It is worth noting that if the policy ${\bm h}$ in the integrated Q-function is replaced with a constant action $a$, the proof of Theorem \ref{thm:known-q-function} will fail. This is the technical reason for us to consider a policy ${\bm h}$  in $Q_{\Delta t}$ in \eqref{equ:def_Q} instead of a constant action $a$.
\end{Remark}

The following result characterizes both the integrated q-function and the value function $J$ associated with a given policy ${\bm \pi}$.

\begin{Theorem}[Characterization of the value function and the integrated q-function] \label{thm:unknown-q-function}
Let ${\bm \pi} \in \Pi$, a continuous function $\hat J: [0, T] \times \Pc_2(\R^d) \to \R$ and a continuous function $\hat q: [0, T] \times \Pc_2(\R^d) \times B_{\delta(\mu)}({\bm \pi})\to \R$ be given.
Then $\hat J$ and $\hat q$ are respectively the value function and the integrated q-function associated with ${\bm \pi}$ if and only if
$\hat J$ and $\hat q$ satisfy
\begin{align} \label{coupled-J-q-terminal-constraint}
\hat J(T, \mu) = \hat g(\mu), \;  \hat q(t, \mu, {\bm \pi}) =0,
\end{align}
and for any $(t, \mu, {\bm h}) \in [0, T] \times \Pc_2(\R^d) \times B_{\delta(\mu)}({\bm \pi})$, the value of
\begin{align}\label{equ:weak_martingale_characterization}
&e^{-\beta s} \hat J(s, \P_{X_s^{\bm h}})- e^{-\beta t} \hat J(t, \mu)+ \int_{t}^s e^{-\beta t'}\big[\hat r(t', \P_{X_{t'}^{\bm h}}, {\bm h}) - \hat q (t', \P_{X_{t'}^{\bm h}}, {\bm h}) {+} {\gamma \E[\mathcal{E}_{\bm h}(t, X_{t'}^{\bm h}, \P_{X_{t'}^{\bm h}})]}\big]dt'
\end{align}
is $0$, where $\{X_s^{\bm h}, t \leq s \leq T\}$ is the solution to \eqref{equ:exploratory_SDE} under the stochastic policy ${\bm h}$ with $\P_{X_t^{\bm h}} = \mu$.
Furthermore, suppose that there exists a function $\hat q_e : [0, T] \times \R^d \times \Pc_2(\R^d) \times \Ac \to \R$ such that $\hat q(t, \mu, {\bm h}) - {\gamma \E_{\xi \sim \mu}[\mathcal{E}_{\bm h}(t, \xi, \mu)]}=  \E_{\mu, {\bm h}} [\hat q_e(t, \xi, \mu, a^{\bm h})]$ for any ${\bm h} \in B_{\delta(\mu)}({\bm \pi})$, 
and it holds further that ${\bm \pi}(a|t, x, \mu) = \frac{\exp\{\frac{1}{\gamma} \hat q_e(t, x, \mu, a)\}}{\int_{\Ac} \exp\{\frac{1}{\gamma} \hat q_e(t, x, \mu, a)\}da}$, then ${\bm \pi}$ is an optimal policy and $\hat J$ is the optimal value function, i.e. $\hat J = J^*$.\\
\end{Theorem}

\textbf{Proof.}  The ``if" part is a direct consequence by following
the same argument as in the proof of Theorem \ref{thm:known-q-function}. We therefore only prove the ``only if" direction.
Note that
\begin{align*}
&e^{-\beta s} \hat J(s, \P_{X_s^{\bm h}})- e^{-\beta t} \hat J(t, \mu)+ \int_{t}^s \Big[\hat r(t', \P_{X_{t'}^{\bm h}}, {\bm h}) -\hat q(t', \P_{X_{t'}^{\bm h}}, {\bm h}) {+}  {\gamma \E[\mathcal{E}_{\bm h}(t, X_{t'}^{\bm h}, \P_{X_{t'}^{\bm h}})]}\Big]dt' = 0.
\end{align*}
From the proof of Theorem \ref{thm:known-q-function}, it holds that
\begin{align*}
&\frac{\partial}{\partial t} \hat J(t, \mu) - \beta \hat J(t, \mu)  + \E\Big[\int_{\Ac} H\big(t, \xi, \mu, a,\partial_\mu \hat J(t, \mu)(\xi), \partial_{x}\partial_\mu \hat J(t, \mu)(\xi)\big){\bm h}(a|t,\xi, \mu)da\Big]\\
& + \gamma \E_{\xi \sim \mu}[\mathcal{E}_{\bm h}(t, \xi, \mu)] - \hat q(t, \mu, {\bm h}) =0. \nonumber
\end{align*}
Letting ${\bm h}= {\bm \pi}$ and combining with $\hat q(t, \mu, {\bm \pi}) = 0$ yields that
\begin{align*}
&\frac{\partial}{\partial t} \hat J(t, \mu) - \beta \hat J(t, \mu)  +
\E\Big[\int_{\Ac} \big\{H\big(t, \xi, \mu, a,\partial_\mu \hat J(t, \mu)(\xi), \partial_{ x}\partial_\mu \hat J(t, \mu)(\xi)\big)\\
& - \gamma{\bm \pi}(a|t, \xi, \mu)\big\}{\bm \pi}(a|t,\xi, \mu)da\Big] =0. \nonumber
\end{align*}
This, together with the terminal condition $\hat J(t, \mu) = \hat g(\mu)$, yields that $\hat J(t, \mu) = J(t, \mu; {\bm \pi})$ by virtue of Feynman-Kac formula \eqref{equ:HJB-coupled-value-function}. Furthermore, based on Theorem \ref{thm:known-q-function}, one has $\hat q(t, \mu, {\bm h}) = q(t, \mu, {\bm h}; {\bm \pi})$.\\

Finally, if ${\bm \pi}(a|t, \mu, x) = \frac{\exp\{\frac{1}{\gamma} \hat q_e(t, x, \mu, a)\}}{\int_{\Ac} \exp\{\frac{1}{\gamma} \hat q_e(t, x, \mu, a)\}da}$, then ${\bm \pi} = \mathcal{I} {\bm \pi}$. This implies that ${\bm \pi}$ is an optimal policy and $\hat J$ is the optimal value function $J^*$.

We end this subsection with the characterization of the optimal value function and optimal essential q-function $q_e$.

\begin{Theorem}[Characterization of the optimal value function and essential q-function]\label{thm:unknown-coupled-optimal-q} Let $\hat J^*: [0, T] \times \Pc_2(\R^d) \to \R$ and $\hat q_e^*: [0, T] \times \R^d \times \Pc_2(\R^d) \times \Ac \to \R$ be continuous functions.
Then $\hat J^*$, $\hat{q}_e^*$ and $\hat {\bm \pi}^*(a|t, x, \mu): = \frac{\exp\{\frac{1}{\gamma} \hat q_e^*(t, x, \mu, a)\}}{\int_{\Ac} \exp\{\frac{1}{\gamma} \hat q_e^*(t, x, \mu, a)\}da}$ are respectively the optimal value function $J^*$, the optimal essential q-function $q_e^*$ and the associated optimal policy ${\bm \pi}^*$ if and only if
\begin{align}\label{optimal-coupled-value-constraint}
\hat J^*(T, \mu) = \hat g(\mu), \; \int_{\R^d} \log \int_{\Ac} \exp\big\{\frac{1}{\gamma} \hat q_e^*(t, x, \mu, a)\big\}da \mu(dx) =0,
\end{align}
and for any $(t, \mu, {\bm h}) \in [0, T] \times \Pc_2(\R^d) \times B_{\delta(\mu)}(\hat{\bm \pi}^*)$, it holds that
\begin{align}\label{thm:weak-martingale-optimal}
&e^{-\beta s} \hat J^*(s, \P_{X_s^{\bm h}})- e^{-\beta t} \hat J^*(t, \mu)+ \int_{t}^s e^{-\beta t'}\Big\{\hat r(t', \P_{X_{t'}^{\bm h}}, {\bm h})\\ &- {\int_{\R^d \times \Ac} \hat q_e^* (t', X_{t'}^{\bm h}, \P_{X_{t'}^{\bm h}}, a){\bm h}(a|t',  X_{t'}^{\bm h}, \P_{X_{t'}^{\bm h}})da \P_{X_{t'}^{\bm h}}(dx)}\Big\}dt' =0 \nonumber.
\end{align}
\end{Theorem}

\begin{Remark}\label{testpolicyexp}
Theorem \ref{thm:unknown-coupled-optimal-q} characterizes the optimal value function and the optimal essential q-function in terms of a weak martingale condition \eqref{thm:weak-martingale-optimal} using all test policies and the consistency condition \eqref{optimal-coupled-value-constraint}, which is the foundation for designing learning algorithms from the social planner's perspective. On one hand, utilizing the martingale condition in \eqref{thm:weak-martingale-optimal} requires us to employ the test policy ${\bm h}$ to generate samples and observations instead of the target policy ${\bm \pi}$, which has a similar flavor of the \textit{off-policy} q-learning for stochastic control by a single agent in \cite{jiazhou2022} that is based on observations of the given behavior policy. On the other hand, we highlight that the feature of mean-field interactions requires us to test all policies ${\bm h}$ in the neighbourhood of the target policy ${\bm \pi}$ to fully characterize the optimal value function and the optimal q-function. The necessity of searching all test policies differs significantly from the standard \textit{off-policy} q-learning in \cite{jiazhou2022} where any given behavior policy together with the consistency condition is sufficient to fully characterize the optimal value function and the optimal q-function. 
\end{Remark}

\begin{Remark}  We emphasize that both the integrated q-function and the essential q-function are crucial in our current setting. The integrated q-function is important from the theoretical perspective while the essential q-function is convenient for implementing the algorithms.

Firstly, recall that in single agent control problems, q-function originates from the time derivative of Q-function in the discrete time framework, which is the main reason that the proposed q-learning algorithms are regarded as the continuous-time counterpart of Q-learning algorithms. As noted in Remark \ref{continuousRM},
our integrated q-function is also the time derivative of the discrete-time integrated Q-function in MFC problems as studied in \cite{GGWX23, CLT23}, however, the discrete-time version of the essential q-function does not exist!  Therefore, without first introducing the integrated q-function, it will be difficult to introduce the definition of the essential q-function. Secondly,  as the essential q-function is not unique (see Remark \ref{continuousRM}), we have to first show that the integrated q-function and the value function can be fully characterized by the weak martingale condition (see Theorem \ref{thm:known-q-function}), which then yields the martingale condition of the essential q-function described in Theorem \ref{thm:unknown-coupled-optimal-q}. For the purpose of the implementation of the algorithm, we can then replace the integrated q-function by the essential q-function in Theorem \ref{thm:unknown-coupled-optimal-q} using the integral representation in \eqref{relationq_dq_cbarq}.
\end{Remark}

\begin{Remark}One relevant work is \cite{FGLPS23}, which investigated the policy gradient algorithm for continuous time MFC problems. It is worth noting that both RL approaches have pros and cons: In their representation of the policy gradient (see (3.3) in Theorem 3.1 in \cite{FGLPS23}), there is an additional term $\Hc$ that can only be computed under the assumption when the model is separable. This term $\Hc$ does not appear in our q-learning approach, which is a main advantage of our q-learning algorithm to cope with MFC problems. However, as a price to pay, we have to utilize infinitely many test polices ${\bm h}$ to fully determine the integrated q-function, which is not needed in \cite{FGLPS23}. When the population distribution cannot be observed, we have also established a comparison between the policy gradient algorithm in \cite{FGLPS23} and the q-learning algorithm from the representative agent's perspective, see proposition 5.6 in \cite{WYY2024}.
\end{Remark}

\section{Algorithms under Continuous Time q-Learning}\label{sec:algorithms}
In this subsection, we assume that the social planner has a simulator for the state distribution and we learn directly the optimal value function and the optimal q-function according to Theorem \ref{thm:unknown-coupled-optimal-q}. First, the parametrized function approximators $J^\theta$ and $q_e^\psi$ are chosen such that the consistency condition \eqref{optimal-coupled-value-constraint} is satisfied that
\begin{align}\label{parametrized-coupled-value-constraint}
J^\theta(T, \mu) = \hat g(\mu)\; \mbox{and} \;\int_{\R^d} \log \int_{\Ac} \exp\big\{\frac{1}{\gamma} q_e^\psi(t, x, \mu, a)\big\}da \mu(dx)=0.
\end{align}
and we have the parameterized policy
$${\bm \pi}^{\psi}(a|t, x, \mu) = \frac{\exp\{\frac{1}{\gamma} q_e^\psi(t, x, \mu, a)\}}{\int_{\Ac} \exp\{\frac{1}{\gamma} q_e^\psi(t, x, \mu, a)da\}}.$$

In what follows, it is assumed that the constant $\int_{\Ac} \exp\{\frac{1}{\gamma} q_e^\psi(t, x, \mu, a)da\}$ is known. We first devise the q-learning algorithm in an offline setting. We can utilize the weak martingale condition \eqref{equ:weak_martingale_characterization} in Theorem \ref{thm:unknown-coupled-optimal-q} to devise the updating rules for parameters by minimizing the loss function robust with respect to all test policies in an offline setting. Denote
\begin{align*}
M_s^{\theta, \psi, {\bm h}} := e^{-\beta s} J^\theta(s, \P_{X_s^{\bm h}})+ \int_{0}^s e^{-\beta t'}\big\{\hat r(t', \P_{X_{t'}^{\bm h}}, {\bm h}) - \int_{\R^d}\int_{\Ac} q_e^\psi(t', x, \P_{X_{t'}^{\bm h}}, a){\bm h}(a|t', x, \P_{X_{t'}^{\bm h}})da \P_{X_{t'}^{\bm h}}\big\}dt',
\end{align*}
where $\P_{X_t^{{\bm h}}}: = \P_{X_{t}^{0, \mu, {\bm h}}}$ is the population state distribution associated with ${\bm h}$ starting from $0$.

According to \eqref{thm:weak-martingale-optimal}, the parameters $\theta$ and $\psi$ will be updated by optimizing mean-square TD error averaged over all test policies that
\begin{align}\label{TD-loss}
\inf_{\theta, \psi}L(\theta, \psi)&: = \frac{1}{2}\inf_{\theta, \psi} \int_{{\bm h} \in B_{\delta(\mu)}({\bm \pi}^\psi)}\big|\dot M_s^{\theta, \psi, {\bm h}} \big|^2ds \nu(d{\bm h}),
\end{align}
where $\dot M_s^{\theta, \psi, {\bm h}}$ is the time derivative of $M_s^{\theta, \psi, {\bm h}}$ and called TD error rate and $\nu$ is a probability measure defined on $B_{\delta(\mu)}({\bm \pi}^\psi) \subset \Pi$ with ${\rm supp}(\nu) = B_{\delta(\mu)}({\bm \pi}^\psi)$. The loss function $L(\theta, \psi)$ has a similar sprit of mean-square temporal difference error in the setting of the deterministic dynamics, see  \cite{Doya2020} and also the discussion in section 3.1 in \cite{JZ22b}.  The infimum of the loss function $\inf_{\theta, \psi} L(\theta, \psi) = 0$ implies that the inside integrand is zero for almost surely $t$ and all ${\bm h}$, which, combined with Theorem \ref{thm:weak-martingale-optimal}, indicates that $J^\theta$ and $q^\psi$ are indeed respectively the optimal value function and the optimal Q-function.

The key issue in the implementation is that we need all test policy ${\bm h}$ to generate sample trajectories to achieve the purpose of learning. In practice, it is impossible to implement all test policies ${\bm h}$ to ensure \eqref{TD-loss} to hold. In this paper, we propose a parametrization method that is based on the fact the test policy has the same parametrized form as the policy ${\bm \pi}^{\psi}$ yet with distinct parameter $\tilde\psi \in B_{\delta}(\psi)$, where $B_\delta (\psi)$ is a ball with radius $\delta$ centered at $\psi$. It is then sufficient to consider the average of the weak martingale loss over $\tilde\psi \in B_\delta(\psi)$.
\begin{align}\label{parametrized-TD-loss}
\inf_{\theta, \psi}\int_{\tilde\psi \in B_\delta(\psi)} \big|\dot M_s^{\theta, \psi, {\bm h}^{\tilde\psi}} \big|^2ds d\tilde\psi.
\end{align}


\textbf{Average-based test policies}:
As the practical implementation is of discrete fashion, we choose a sequence of test policies $\bm h^{\tilde \psi^1}$, $\bm h^{\tilde\psi^2}$, $\ldots,$ $\bm h^{\tilde\psi^M}$ using the parameterized policy ${\bm \pi}^{\psi}$. 
More precisely, at the beginning of each episode $j$, we use $\psi$ to randomly generate $M$ parameters $\tilde\psi^1, \ldots, \tilde\psi^M$, which corresponds to $M$ test policies ${\bm h}^{\tilde\psi^1}, \ldots, {\bm h}^{\tilde\psi^M}$. For example, $\tilde\psi^1, \ldots, \tilde\psi^M$ are i.i.d. drawn from $\psi \cdot \mathcal{U}([p(j), q(j)])$, where $\mathcal{U}([p(j), q(j)])$ is uniform distribution on $[p(j), q(j)]$. 

In this case, we discretize $[0, T]$ on the grid $\{t_k = k\Delta t, k=0, 1, \ldots, K\}$ and hence the weak martingale loss function is averaged over all test policies
\begin{align}\label{discrete-parametrized-TD-loss}
&\frac{1}{2M}\sum_{m=1}^M \sum_{k=0}^{K - 1} \big|\delta_k^{\theta, \psi, {\bm h}^{\tilde\psi^m}}\big|^2\Delta t,
\end{align}
where $\P_{X_t^{{\bm h}^{\tilde\psi^m}}}: = \P_{X_{t}^{{0, \mu, {\bm h}^{\tilde\psi^m}}}}$ is the population state distribution associated with ${\bm h}^{\tilde\psi^m}$, $1 \leq m \leq M$ starting from $0$, and $\delta_k^{\theta, \psi, {\bm h}^{\tilde\psi^m}}$ is given by
\begin{align*}
\delta_k^{\theta, \psi, {\bm h}^{\tilde\psi^m}} =& e^{-\beta t_k}\Big(\frac{J^\theta(t_{k+1}, \mu_{t_{k+1}}^m) - J^\theta(t_{k}, \mu_{t_{k}}^m)}{\Delta t} + \hat r_{t_k}^m  - \beta J^\theta(t_k, \mu_{t_{k}}^m)\\
& - \int_{\R^d \times \Ac} q_e^\psi(t_k, x, \mu_{t_k}^m, a){\bm h}^{\tilde\psi^m}(a|t_k, x, \mu_{t_k}^m)da\mu_{t_k}^m(dx)\Big).
\end{align*}
It is shown in Theorem 5.4 in \cite{WYY2024} that as $M \to +\infty$ and $\Delta t \to 0$, the solution of \eqref{discrete-parametrized-TD-loss} converges to that of \eqref{parametrized-TD-loss}.

We then apply the vanilla gradient descent to update $\theta$ and $\psi$:
\begin{align*}
\theta \leftarrow \theta - \alpha_\theta \frac{1}{M}\sum_{m=1}^M \Delta^m \theta, \; \psi \leftarrow \psi - \alpha_{\psi} \frac{1}{M}\sum_{m=1}^M \Delta^m \psi,
\end{align*}
where $\mu_{t_k}^{m}$ and $\hat r_{t_k}^m$, $0 \leq k \leq K-1$ are observed state distribution and observed aggregated reward associated with the test policy ${\bm h}^{\tilde\psi^m}$ , and $\alpha_\theta$ and $\alpha_{\psi}$ are learning rates and
\begin{align} \label{method1-expression-G-tk-T}
G_{t_k}^m & = e^{-2\beta t_k}\Big({J^\theta(t_{k+1}, \mu_{t_{k+1}}^m) - J^\theta(t_{k}, \mu_{t_{k}}^m)}+ \big(\hat r_{t_k}^m  - \beta J^\theta(t_k, \mu_{t_{k}}^m)\\
& - \int_{\R^d \times \Ac} q_e^\psi(t_k, x, \mu_{t_k}^m, a){\bm h}^{\tilde\psi^m}(a|t_k, x, \mu_{t_k}^m)da\mu_{t_k}^m(dx)\big)\Delta t\Big), \nonumber\\
\Delta^m \theta & = \sum_{k=0}^{K-1}G_{t_k}^m \Big(\frac{1}{\Delta t}\big(\frac{\partial J^\theta}{\partial \theta}(t_{k+1}, \mu_{t_{k+1}}^m) - \frac{\partial J^\theta}{\partial \theta}(t_{k}, \mu_{t_{k}}^m)\big) - \beta J^\theta(t_k, \mu_{t_k}^m) \Big),\\
\Delta^m \psi &= -\sum_{k=0}^{K-1} G_{t_k}^m \int_{\R^d \times \Ac} \frac{\partial q_e^{\psi}}{\partial \psi}(t_i, x, \mu_{t_i}^m, a){\bm h}^{\tilde\psi^m}(a|t_i, x, \mu_{t_i}^m) da \mu_{t_i}^m(dx). \label{method1-expression-Deltam-psi}
\end{align}

The pseudo-code is described in Algorithm \ref{algo:offline episodic ml-method1}. 

\begin{Remark}
Generally speaking, it is computationally costly to evaluate the integrals in  \eqref{method1-expression-G-tk-T}-\eqref{method1-expression-Deltam-psi}. However, in numerical examples in Section \ref{sec:application}, we can leverage the model structure to explicitly compute these integrals. In particular, in the first example in subsection \ref{sec:MV}, $q_e^\psi$ is quadratic in $x$ and $a$, and the parameterized test policy ${\bm h}^{\tilde\psi}$ is taken as the normal distribution so that the integrals become functions of expectation and variance of the population distribution $\mu$. In general cases, we may have to compute these integrals by a Monte Carlo approach or the approximation of $q_e^\psi$ using simple functions, such as polynomials of $x$ and $a$.
\end{Remark}

As the social planner has access to the full information of the population distribution, the population distribution is generated by a simulator $(\mu',\hat r) = \textit{Environment}_{\Delta t}(t, \mu, {\bm h})$ in Algorithms \ref{algo:offline episodic ml-method1}-\ref{algo:online episodic ml} that is based on the time discretized version of Fokker-Planck equation. More precisely, 
consider the time discretized SDE
\begin{align*}
X_{t_{k+1}} &\approx X_{t_k} + b(t_k, X_{t_k}, \mu_{t_{k}}, a^{{\bm h}}) \Delta t + \sigma(t_k, X_{t_k}, \mu_{t_k}, a_t^{{\bm h}}) (W_{t_{k+1}} - W_{t_k}), \; a^{{\bm h}} \sim {\bm h}(\cdot|t_k,  X_{t_k}, \mu_{t_k}).
\end{align*}
Denote the probability transition function of by $p(dx'|t, X_{t_k}, \mu_{t_k}, a^{{\bm h}}): = \P(X_{t_{k+1}}\in dx'| X_{t_k})$,
then $p$ is a normal distribution $\Nc\big(X_{t_k} + b(t_k, X_{t_k}, \mu_{t_k}, a^{{\bm h}}) \Delta t, \sigma\sigma\trans(t_k, X_{t_k}, \mu_{t_k}, a^{{\bm h}})\Delta t\big)$. By the law of total probability
\begin{align}\label{equ:discrete_FP}
\mu_{t_{k+1}}(dx') = \int_{\R^d} \int_{\Ac} p(dx'|t, x, \mu_{t_k}, a){\bm h}(a|t, x, \mu_{t_k}) da \mu_{t_k}(dx).
\end{align}
\eqref{equ:discrete_FP} implies the evolution of population state distribution ${\mu_{t_k}}$ over the time.

\begin{Remark}
In practice, instead of computing \eqref{equ:discrete_FP}, we assume that the social planner has access to an environment simulator that updates moments of the population distribution. Precisely, this is represented by
$$(\bar\mu_{t_{k+1}}, \bar\mu_{2, t_{k+1}}, \ldots, \bar\mu_{m, t_{k+1}}, \hat r) = \textit{Environment}_{\Delta t}(t_k, \bar\mu_{t_{k}}, \bar\mu_{2, t_{k}}, \ldots, \bar\mu_{m, t_{k}}, {\bm h}),\;\; m \in \N^*,$$
where $\bar\mu_{t, m} = \int_{\R^d} x^m \mu_t(dx)$ denotes the $m$-th moment of $\mu$. For example, moments up to second order are sufficient for the linear quadratic framework. See \eqref{MV:barmu-update}-\eqref{MV:var-update} for the simulator of the mean-variance example and \eqref{R&Dlog-barmu-update} for mean-field optimal consumption problem.
\end{Remark}

\begin{Remark} The information structure (i.e. what an agent can observe) plays a significant role in the reinforcement learning of MFC problems. In this study, we consider MFC from the perspective of the social planner, who can observe {\it macroscopic} quantities, including population distributions $\mu_t$ and aggregated rewards $\hat r_t$. In contrast, \cite{WYY2024} studies MFC from the viewpoint of the representative agent, who lacks the access to population distributions $\mu_t$ and can only observe his own states $x_t$ and rewards $r_t$. It is assumed therein that the representative agent updates the population distributions based on his observed states. Consequently, different information structures will lead to distinct q-functions with different martingale characterizations and environment simulators.
\end{Remark}

We can also similarly devise the q-learning algorithm in an online-setting where the parameters are updated in real-time. Similar to the previous algorithm, we can minimize the loss function over all average-based test policies $\{{\bm h}^{\tilde\psi^1}, \ldots, {\bm h}^{\tilde\psi^M}\}$ between $t_k$ and $t_{k+1}$ that
\begin{align*}
&\frac{1}{2M}\sum_{m=1}^M \Big|J^\theta(t_{k+1}, \P_{X_{t_{k+1}}^{{\bm h}^{\tilde\psi^m}}})- J^\theta(t_k,  \P_{X_{t_{k}}^{{\bm h}^{\tilde\psi^m}}})+ \Big(\hat r(t_k, \P_{X_{t_k}^{{\bm h}^{\tilde\psi^m}}}, {\bm h}^{\tilde\psi^m})\\
& \;\;-\int_{\R^d} \int_{\Ac} q_e^{\psi}(t_k, x, \P_{X_{t_{k}}^{{\bm h}^{\tilde\psi^m}}}, a){\bm h}^{\tilde\psi^m}(a|t_k, x, \P_{X_{t_{k}}^{{\bm h}^{\tilde\psi^m}}}) \P_{X_{t_{k}}^{{\bm h}^{\tilde\psi^m}}}(dx) - \beta J^\theta(t_k,  \P_{X_{t_{k}}^{{\bm h}^{\tilde\psi^m}}}) \Big)\Delta t\Big|^2.
\end{align*}
We can then apply gradient descent (GD) to the above loss function after the time discretization of $[0, T]$ on the grids $\{t_k = k \Delta t, k=0, \ldots, K-1\}$ and obtain the updating rules for $\theta$ and $\psi$ that
\begin{align*}
\theta \leftarrow \theta - \alpha_\theta \frac{1}{M}\sum_{m=1}^M \Delta^m \theta, \; \psi\leftarrow \psi - \alpha_{\psi} \frac{1}{M} \sum_{m=1}^M \Delta^m \psi,
\end{align*}
where $\mu_{t_k}^{m}, 0 \leq k \leq K-1$ and $\hat r_{t_k}^m, 0 \leq k \leq K-1$ are observed state distribution and reward associated to the test policy ${\bm h}^{\tilde\psi^m}$ , and $\alpha_\theta$ and $\alpha_{\psi}$ are learning rates and
\begin{align}\label{online-method2-expression-delta-k-m}
\delta_{k}^m &=J^\theta(t_{k+1}, \mu_{t_{k+1}}^m)- J^\theta(t_k,  \mu_{t_k}^m)+ \Big(\hat r_{t_k} - \beta J^\theta(t_k,  \mu_{t_k}^m) \\
 &-\int_{\R^d} \int_{\Ac} q_e^{\psi}(t_k, x, \mu_{t_k}^m, a){\bm h}^{\tilde\psi^m}(a|t_k, x, \mu_{t_k}^m) \mu_{t_k}^m(dx) \Big)\Delta t, \nonumber\\
 \Delta^m \theta & = \delta_{k}^m  \big(\frac{\partial J^\theta}{\partial \theta}(t_{k+1}, \mu_{k+1}^m) - \frac{\partial J^\theta}{\partial \theta}(t_k,  \mu_{t_k}^m) - \beta J^\theta(t_k,  \mu_{t_k}^m) \Delta t\big),\\
\Delta^m \psi &= -\delta_{k}^m \int_{\R^d} \int_{\Ac}\frac{\partial q_e^{\psi}}{\partial \psi}(t_k, x, \mu_{t_k}^m, a) {\bm h}^{\tilde\psi^m}(a|t, x, \mu_{t_k}^m) \mu_{t_k}^m(dx) \Delta t. \label{online-method2-expression-deltam-psi}
 \end{align}
The pseudo-code is described in Algorithm \ref{algo:online episodic ml}. We remark that the dynamic is deterministic from the perspective of the social planner and thus the updating rule has a similar spirit as Doya's Temporal Difference (TD) algorithm for deterministic dynamics \cite{Doya2020}.

\begin{Remark}
For Algorithms \ref{algo:offline episodic ml-method1} and \ref{algo:online episodic ml}, we again emphasize that the test policy ${\bm h}$ is used to the environment simulator instead of the target policy $\bm \pi$, which makes our algorithms similar to the so-called off-policy in conventional Q-learning. However, the feature of mean-field interactions requires us to explore all test policies to meet the weak martingale condition, and hence makes our algorithms different from the standard off-policy learning. The parameters in the policy (or q-function) are used and updated in the learning procedure but the resulting policy ${\bm \pi}$ does not participate in the procedure directly. All samples and observations are based on our chosen test policy ${\bm h}$, which can be adaptively updated by our proposed two different methods.
\end{Remark}

\ \\
\begin{algorithm}[hbtp]
\caption{Offline q-Learning Algorithm}
\textbf{Inputs}: initial state distribution $\mu_0$,  horizon $T$, time step $\Delta t$, number of mesh grids $K$, number of test policies $M$, initial learning rates $\alpha_{\theta}$ and $\alpha_{\psi}$, functional forms of parameterized  value function $J^{\theta}(\cdot,\cdot)$ and $q_e^{\psi}(\cdot,\cdot,\cdot, \cdot)$ satisfying \eqref{optimal-coupled-value-constraint} and temperature parameter $\gamma$.

\textbf{Required program}: Moments simulator or environment simulator $(\mu',\hat r) = \textit{Environment}_{\Delta t}(t, \mu, {\bm h}^{\tilde\psi})$ that takes current time--state distribution pair $(t, \mu)$ and the test policy ${\bm h}^{\tilde\psi}$ as inputs and generates state distribution $\mu'$ at time $t+\Delta t$ and the aggregated reward $\hat r$ at time $t$ as outputs.

\textbf{Learning procedure}:
\begin{algorithmic}
\STATE Initialize $\theta$ and $\psi$.
\FOR{episode $j=1$ \TO $N$}
\STATE{Observe  the initial state distribution $\mu_0$ and store $\mu_{t_k}^m \leftarrow  \mu_0$.}
\FOR{$m=1$ \TO $M$}
\STATE{Draw $\tilde\psi^m$ from $\psi \cdot \mathcal{U}([p(j), q(j)])$ and set the test policy ${\bm h}^{\tilde\psi^m}$.}
\STATE{Initialize $k = 0$.
\WHILE{$k < K$} \STATE{
Apply the test policy ${{\bm h}}^{\tilde\psi^m}$ to the environment simulator $(\mu,\hat r) = Environment_{\Delta t}(t_k, \mu_{t_k}^m, {\bm h}^{\tilde\psi^m})$, and observe the new state distribution $\mu$ and the aggregated reward $\hat r$ as output. Store $\mu_{t_{k+1}}^m \leftarrow \mu$ and $\hat r_{t_k}^m \leftarrow \hat r$.

Update $k \leftarrow k + 1$.
}
\ENDWHILE	

For every $k=0,1,\cdots,K-1$, compute $G_{t_k: T}^m$, $\Delta^m \theta$ and $\Delta^m \psi$ according to \eqref{method1-expression-G-tk-T}-\eqref{method1-expression-Deltam-psi}.
}
\ENDFOR

Update $\theta$ and $\psi$ by $\theta \leftarrow \theta - \alpha_\theta \frac{1}{M} \sum_{m=1}^M \Delta^m \theta, \psi \leftarrow \psi - \alpha_{\psi} \frac{1}{M} \sum_{m=1}^M \Delta^m \psi$.

\ENDFOR
\end{algorithmic}
\label{algo:offline episodic ml-method1}
\end{algorithm}


\begin{algorithm}[hbtp]
\caption{Online q-Learning Algorithm}
\textbf{Inputs}: initial state distribution $\mu_0$,  horizon $T$, time step $\Delta t$, number of mesh grids $K$,  number of behavior policies $M$, initial learning rates $\alpha_{\theta}$ and $\alpha_{\psi}$, functional forms of parameterized  value function $J^{\theta}(\cdot,\cdot)$ and $q_e^{\psi}(\cdot,\cdot,\cdot, \cdot)$ satisfying \eqref{optimal-coupled-value-constraint} and temperature parameter $\gamma$.

\textbf{Required program}: Moments simulator or environment simulator $(\mu',\hat r) = \textit{Environment}_{\Delta t}(t, \mu, {\bm h}^{\tilde\psi})$ that takes current time--state distribution pair $(t, \mu)$ and the test policy ${\bm h}^{\tilde\psi}$ as inputs and generates the state distribution $\mu'$ at time $t+\Delta t$ and the aggregated reward $\hat r$ at time $t$ as outputs.

\textbf{Learning procedure}:
\begin{algorithmic}
\STATE Initialize $\theta$ and $\psi$.
\FOR{episode $j=1$ \TO $N$}
\STATE{Initialize $k = 0$. Observe the initial state distribution $\mu_0$ and store $\mu_{t_k}^m \leftarrow  \mu_0$, $m=1, \ldots, M$.
\WHILE{$k < K$}
\FOR{$m=1$ \TO $M$}
\STATE{Draw $\tilde\psi^m$ from $\psi \cdot \mathcal{U}([p(j), q(j)])$ and set the test policy ${\bm h}^{\tilde\psi^m}$}.
Apply the test policy ${{\bm h}}^{\tilde\psi^m}$ to environment simulator $(\mu,\hat r) = Environment_{\Delta t}(t_k, \mu_{t_k}^m, {\bm h}^{\tilde\psi^m})$, and observe  new state distribution $\mu$ and the aggregated reward $\hat r$ as output. Store $\mu_{t_{k+1}}^m \leftarrow \mu$ and $\hat r_{t_k}^m \leftarrow \hat r$.

Compute $\delta_k^m, \Delta^m \theta$ and $\Delta^m \psi$ according to \eqref{online-method2-expression-delta-k-m}-\eqref{online-method2-expression-deltam-psi}.

\ENDFOR

Update $\theta$ and $\psi$ by
$\theta \leftarrow \theta -\alpha_\theta \frac{1}{M} \sum_{m=1}^M \Delta^m \theta, \psi \leftarrow \psi - \alpha_{\psi} \frac{1}{M}\sum_{m=1}^M\Delta^m \psi$.

Update $k \leftarrow k + 1$.

\ENDWHILE
}

\ENDFOR
\end{algorithmic}
\label{algo:online episodic ml}
\end{algorithm}

\newpage
\section{Financial Applications}\label{sec:application}

\subsection{Mean-Variance Portfolio Optimization}\label{sec:MV}
Let us consider the wealth process that satisfies the SDE that
\begin{align}\label{equ:MV_exploratory_average_SDE}
dX_s^{\bm \pi} = a_s^{\bm \pi}\big(bds +\sigma dW_s\big), a_s^{\bm \pi} \sim {\bm \pi}(\cdot|s, X_s^{\bm \pi}, \P_{X_s^{\bm \pi}}),\; s \geq t, \; X_0 = x,
\end{align}
where $a_s^{\bm \pi}$ is the wealth amount invested in the risky asset at time $s$, $b$ is the excess return and $\sigma$ is the volatility. The learning mean-variance portfolio optimization problem with entropy regularizer is defined by
\begin{align*}
J(t, \mu; {\bm \pi}) = \E[X_T^{\bm \pi}] -\lambda {\rm Var}(X_T^{\bm \pi}) - \gamma  \E\biggl[\int_t^T \int_{\R}\log {\bm \pi}(a|s, X_s^{\bm \pi}, \P_{X_s^{\bm \pi}}) {\bm \pi}(a|s, X_s^{\bm \pi}, \P_{X_s^{\bm \pi}}) da \biggl].
\end{align*}
The Hamiltonian operator is given by $H(t, x, \mu, a, p, q) = ba p + \frac{1}{2} \sigma^2 a^2  q$. It then follows that
\begin{align*}
\int_{\R}\exp\Big\{\frac{1}{\gamma} H(t, x, \mu, a, p, q)\Big\} da &=\exp\big(-\frac{b^2 p^2}{2\sigma^2 q \gamma}\big)  \int_{\R} \exp \Big(\frac{1}{2\gamma} q \sigma^2 (a + \frac{bp}{q\sigma^2})^2 \Big)da\\
& = \exp\big(-\frac{b^2 p^2}{2\gamma \sigma^2 q}\big)  \sqrt{-\frac{2\pi\gamma}{q\sigma^2}}.
\end{align*}
As $J^*(t, \mu)$ satisfies \eqref{equ:exploratory_HJB}, we have $J^*(T, \mu) = -\lambda {\rm Var}(\mu) + \bar\mu$ and the exploratory HJB equation is given by
\begin{align} \label{equ:MVexample_expoloratory}
& \frac{\partial J^*}{\partial t}(t, \mu) + \gamma \int_{\R} \log \int_{\R} \exp\Big\{\frac{ba}{\gamma}\partial_\mu J^*(t, \mu)({x}) +\frac{\sigma^2a^2}{2\gamma} \partial_{x}\partial_\mu J^*(t, \mu)({x})\Big\} da \mu(d{x})\\
 = & \frac{\partial J^*}{\partial t}(t, \mu) - \frac{b^2}{\sigma^2} \int_{\R}\frac{\Big|\partial_\mu J^*(t, \mu)({x})\Big|^2}{2\partial_{x}\partial_\mu J^*(t, \mu)({x})}\mu(d{x}) + \frac{\gamma}{2} \int_{\R}\log \frac{2\pi \gamma}{-\sigma^2 \partial_{x}\partial_\mu J^*(t, \mu)({x})} \mu(d{x})=0. \nonumber
\end{align}
Denote ${\rm Var}(\mu): = \int_{\R} (x - \bar\mu)^2 \mu(dx), \; \bar\mu := \int_{\R} x\mu(dx)$. We conjecture that $J^*(t, \mu)$ satisfies the quadratic form
\begin{align}
J^*(t, \mu) = A(t){\rm Var}(\mu) +  C(t)\bar \mu  +D(t).\label{conject-1}
\end{align}
It then holds that
\begin{align*}
\frac{\partial J^*}{\partial t}(t, \mu) &= \dot A(t){\rm Var}(\mu) + \dot C(t)\bar \mu  + \dot D(t),\\
\partial_\mu J^*(t, \mu)({x}) &= 2A(t) ({x} -\bar\mu) + C(t),\quad\quad \partial_{{x}}\partial_\mu J^*(t, \mu)({x}) = 2A(t).
\end{align*}
Plugging these into the exploratory HJB equation \eqref{equ:MVexample_expoloratory}, we get that
\begin{align*}
&\Big[\dot A(t) - \frac{b^2}{\sigma^2}A(t)\Big]{\rm Var}(\mu) + \dot C(t) \bar\mu + \Big[\dot D(t) - \frac{b^2}{\sigma^2} \frac{C(t)^2}{4A(t)} + \frac{\gamma}{2} \log \frac{\pi \gamma}{-\sigma^2 A(t)}\Big] =0.
\end{align*}
By the terminal conditions $A(T) = -\lambda$, $C(T) = 1$ and $D(T) =0$, we can obtain the explicit solution of the ODEs that
\begin{align*}
&A(t) =-\lambda \exp\big(\frac{b^2}{\sigma^2}(t-T)\big), \; C(t) =1,\\
& D(t) = \frac{\gamma b^2}{4\sigma^2} (t- T)^2 - (t -T) \frac{\gamma}{2} \log \frac{\pi\gamma}{\sigma^2\lambda} + \frac{1}{4\lambda} \exp\big(-\frac{b^2}{\sigma^2}(t-T)\big) - \frac{1}{4\lambda}.
\end{align*}
Then, we can take the essential q-function $q_e^*$ as
\begin{align*}
q_e^*(t, x, \mu, a) & =-\lambda\sigma^2 \exp\big(\frac{b^2}{\sigma^2}(t-T)\big) \Big(a + \frac{b}{\sigma^2} (x - \bar\mu) - \frac{b}{2\lambda\sigma^2} \exp(-\frac{b^2}{\sigma^2}(t-T))\Big)^2\\
& \;\;\;- \frac{b^2 \lambda}{\sigma^2} \exp(\frac{b^2}{\sigma^2}(t-T)) \big[(x- \bar\mu)^2 - {\rm Var}(\mu)\big] - \frac{b^2}{\sigma^2} (x - \bar\mu)- \frac{\gamma}{2} \log \frac{\pi \gamma}{\sigma^2 \lambda} + \frac{\gamma b^2}{2\sigma^2}(t -T).
\end{align*}
It then follows 
that the optimal policy ${\bm \pi}^*$ is
\begin{align} \label{mean-variance-pi*}
{\bm \pi}^*(\cdot|t, x, \mu) = \Nc\Big(-\frac{b}{\sigma^2} \big(x - \bar\mu - \frac{1}{2\lambda}\exp(-\frac{b^2}{\sigma^2}(t-T))\big), \frac{\gamma}{2\lambda \sigma^2}\exp(-\frac{b^2}{\sigma^2}(t-T))\Big).
\end{align}

When model parameters $b$, $\sigma$ and $\lambda$ are unknown, we can derive the parameterized functions $J^\theta$ and $q_e^\psi$ by
\begin{align}
J^\theta(t, \mu) &= -\frac{1}{4\theta_3} \exp(\theta_1(t-T)) {\rm Var}(\mu) +  \bar\mu + \frac{\gamma \theta_1}{4}(t-T)^2 + \theta_2 (t-T)\\
&\; + \theta_3 \exp(-\theta_1(t-T)) - \theta_3, \nonumber\\
q_e^\psi(t, x, \mu, a) &= -\frac{\exp(\psi_1 + \psi_2(t- T))}{2} \big(a + \psi_3(x - \bar\mu) + \psi_4 \exp(-\psi_2(t-T))\big)^2 - \frac{\gamma}{2}\log (2\pi \gamma) + \frac{\gamma}{2}{\psi_1} \nonumber\\
& \;+ \frac{\psi_2 \gamma}{2}(t - T) - \psi_2(x-\bar\mu) + \psi_5 \exp(\psi_2(t- T)))\big((x - \bar\mu)^2 - {\rm Var}(\mu)\big), \label{MV:para-essential-q}
\end{align}
where $\theta = (\theta_1, \theta_2, \theta_3)\trans \in \R^3$, $\psi = (\psi_1, \ldots, \psi_5) \in \R^5$. Note that the above parameterizations satisfy the constraints in \eqref{parametrized-coupled-value-constraint}. It then follows that
${\bm \pi}^\psi(\cdot|t, x, \mu) = \Nc( -\psi_3(x - \bar\mu) - \psi_4 \exp(-\psi_2(t-T))\big), \gamma\exp(-\psi_1-\psi_2(t-T)))$. Note that $\psi_5$ does not make any contribution to learn ${\bm \pi}^{\psi}$ and is redundant to be learnt by the q-learning algorithm.

The following simulator will be used in Algorithms \ref{algo:offline episodic ml-method1}-\ref{algo:online episodic ml} to generate sample trajectories.

\noindent {\bf Simulator} Under the policy ${\bm h}^{\tilde \psi}$, the equation \eqref{equ:MV_exploratory_average_SDE} becomes
\begin{align*}
d\tilde X_t &= b \int_{\R} a {\bm h}^{\tilde\psi}(a|t, \tilde X_t, \P_{{\tilde X}_t})da dt  + \sigma \sqrt{a^2 {\bm h}^{\tilde\psi}(a|t, \tilde X_t , \P_{\tilde X_t})da}dW_t\\
& = - b \Big(\tilde\psi_3\big(\tilde X_t - \E[ \tilde X_t]\big) +  \tilde\psi_4 \exp(-\tilde {\psi}_2(t-T) \Big) dt\\
& \;\;\;+ \sigma \sqrt{ \Big(\tilde\psi_3(\tilde X_t - \E[\tilde X_t]) + \tilde\psi_4\exp(-\tilde\psi_2(t-T)) \Big)^2 + \gamma\exp(-\tilde\psi_1-\tilde\psi_2(t-T))} dW_t.
\end{align*}
First we calculate the mean of $\tilde X_t$ that $\bar \mu_t = \bar{\P}_{\tilde X_t^{{\bm h}^\psi}}$ by taking the expectation on both sides of the above SDE, which yields that
$d \E[\tilde X_t] = - b \tilde\psi_4 \exp(-\tilde\psi_2(t-T)) dt$.
We thus deduce that
\begin{align}\label{MV:barmu-update}
\bar\mu_{t_{k+1}} \simeq \bar\mu_{t_k}   - b\tilde\psi_4 \exp(-\tilde\psi_2(t_k-T)) \Delta t.
\end{align}

We next compute the variance of $\tilde X_t$: ${\rm Var}({\P}_{\tilde X_t}) = {\rm Var}(\tilde X_t)$ by applying It\^o's formula to $\big(\tilde X_t - \E[\tilde X_t]\big)^2$ and then taking expectation that
\begin{align*}
d {\rm Var}(\tilde X_t) = \Big((\sigma^2 \tilde\psi_3^2 - 2b\tilde\psi_3) {\rm Var}(\tilde X_t)  + \sigma^2 \tilde\psi_4^2  \exp(-2\tilde\psi_2(t-T)) + \gamma\exp(-\tilde\psi_1-\tilde\psi_2(t-T))\Big) dt.
\end{align*}
The Euler approximation of ${\rm Var}(\tilde X_t)$ is given by
\begin{align}\label{MV:var-update}
{\rm Var}(\tilde X_{t_{k+1}}) &= {\rm Var}(\tilde X_{t_k}) + \Big((\sigma^2 \tilde\psi_3^2 - 2b\tilde\psi_3) {\rm Var}(\tilde X_{t_k})  + \sigma^2 \tilde\psi_4^2\exp(-2\tilde\psi_2(t_k-T))\\
& \;\;+ \gamma \sigma^2\exp(-\tilde\psi_1-\tilde\psi_2(t_k-T))\Big)\Delta t. \nonumber
\end{align}
Note that the simulator in terms of $(\E[\tilde X_{t_k}^{{\bm h}^{\tilde\psi}}], {\rm Var}(\tilde X_{t_k}^{{\bm h}^{\tilde\psi}}))$ is deterministic. The aggregate reward is simulated according to $\hat r_{t_k} = 0$ and $\hat r_{t_K} = \E[\tilde X_T] - \lambda{\rm Var}(\tilde X_T)$.


We first set the coefficients of the simulator to $T=1, b=0.25, \sigma =0.5, {\lambda = 1.5}, \; \beta = 0$. Nest, we set the known model parameters as: $\gamma = 0.5$, $\beta = 0$, the time step $\Delta t = 0.05$, the number of episodes $N = 2500$, the number of test policies $M = 10$,  the lower bound of uniform distribution $p(j) = 0$, the upper bound of uniform distribution $q(j) = \frac{2}{j^{0.25}}$. We set the initialization of $\bar\mu_0^m \sim \Nc(0, 1 )$ and  ${\rm Var}^m(\mu_0) \sim \Uc([0, 1])$ respectively, and choose the initialization of $\theta = (-0.5, 0.5, 0.5)\trans$, $\psi = (0.5, -0.5, 1.5, -0.5)\trans$. The learning rates $(\alpha_\theta, \alpha_\psi)$ are chosen by
\begin{align*}
 \alpha_\theta(j) &= (\frac{0.015}{j^{0.22}}, \frac{0.01}{j^{0.1}}, \frac{0.025}{j^{0.11}}),\\
 \alpha_\psi(j) &= (\frac{0.035}{j^{0.09}}, \frac{0.11}{j^{0.1}}, \frac{0.02}{j^{0.2}}, \frac{0.01}{j^{0.15}}).
 \end{align*}

\begin{figure}[tbhp]
\centering
\includegraphics[width=0.4\linewidth]{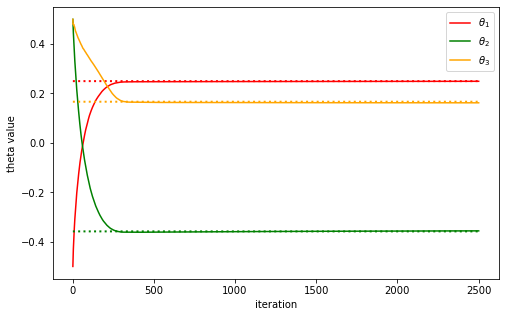}
\includegraphics[width=0.4\linewidth]{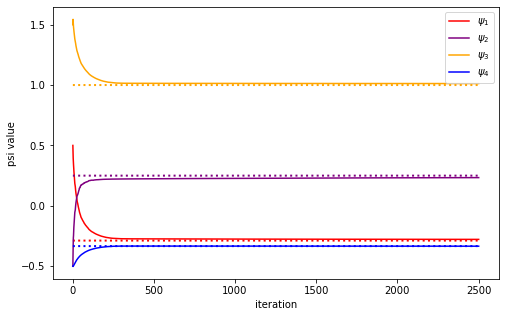}
\includegraphics[width=0.4\linewidth]{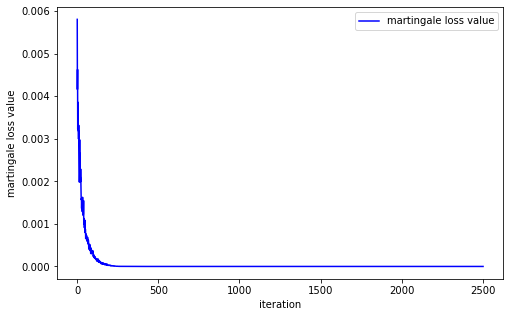}
\includegraphics[width=0.4\linewidth]{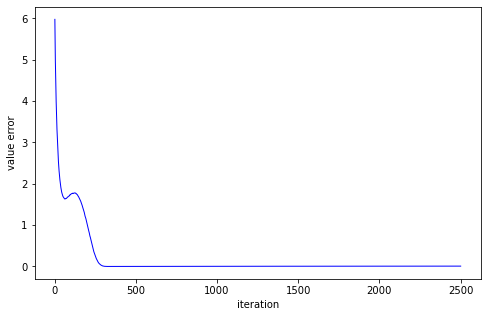}
\caption{\footnotesize {\bf Convergence of value function and q-function for Algorithm \ref{algo:offline episodic ml-method1}}. Paths of learnt parameters for value function (top left) and paths of learnt parameters for q-function (top right) vs optimal parameters shown in the dashed line. The change of the weak martingale loss value over iterations (bottom left) and the change of $L^2$- error over iterations (bottom right) along a trajectory $(\bar\mu_{t_k}, {\rm Var}(\mu)_{t_k})_k$ with $\bar\mu_0 =0$ and ${\rm Var}(\mu)_0 =0.5$ controlled by the learnt policy and by the optimal policy.}
\label{fig:algorithm1-mean-variance}
\end{figure}

Based on the offline q-learning Algorithm \ref{algo:offline episodic ml-method1}, we plot in Figure \ref{fig:algorithm1-mean-variance} the numerical results on the convergence of parameters $\theta$ and $\psi$ for the optimal value function and the optimal essential q-function $q_e$, and summarize the learnt parameters for the optimal value function and the optimal essential q-function $q_e$ in Table \ref{table:parameters-mean-variance}.

{\color{blue}
\begin{table}
 \caption{}
 \centering
 \begin{tabular}{cccccccc}
 \hline
 & $\theta_1$ & $\theta_2$ & $\theta_3$ & $\psi_1$ & $\psi_2$ & $\psi_3$ & $\psi_4$\\
 \hline
 true value & 0.25  & -0.358 & 0.1667 & -0.288 &   0.25 &   1  & -0.333\\
 \hline
 learnt by Algorithm \ref{algo:offline episodic ml-method1} & 0.249 & -0.356 & 0.162 & -0.288 & 0.234 & 1.012 & -0.334\\
 \hline
 \end{tabular}
 \label{table:parameters-mean-variance}
 \end{table}
}

\subsection{Mean-Field Optimal Consumption Problem}
We consider the mean-field R$\&$D project process that is governed by the McKean-Vlasov SDE
\begin{align}\label{equ:mean-field-consumption-dynamics}
d  X_s = a_s b\E[X_s] ds + \sigma \E[ X_s] dW_s - c_s ds, (a_s, c_s) \sim {\bm \pi}(\cdot|s,  X_s, \P_{X_s}), \; X_t = x >0,
\end{align}
with $b>0$ and $\sigma>0$. Here, $a_s$ is the drift control of the R$\&$D investment and $c_s > 0$ is the consumption rate.
Let us consider the problem of expected utility on consumption subjecting to the quadratic cost of control in the following form that
\begin{align*}
J(t, \mu; \{a_s\}_{s\geq t}, \{c_s\}_{s \geq t}) = \E\biggl[\int_t^T e^{-\beta (s-t)}U(c_s)ds -\int_t^T e^{-\beta (s-t)}a_s^2ds\biggl]
\end{align*}
with the logrithmic utility function $U(x) = \log  x$. The exploratory control-consumption performance after the entropy regularizer is defined by
{\small
\begin{align*}
J(t, \mu; {\bm \pi}) &= \E\biggl[\int_t^T \int_{\R \times \R_+} \biggl\{e^{-\beta(s-t)}\Big(U(c) -e^{-\beta(s-t)}a^2 -\gamma \log {\bm \pi}(a, c|s,  X_s^{\bm \pi}, \P_{ X^{\bm \pi}_s})\Big)\biggl\} {\bm \pi}(a, c|s, X_s^{\bm \pi}, \P_{X^{\bm \pi}_s}) dadc ds\biggl].
\end{align*}}The Hamiltonian operator is given by
\begin{align*}
H(t, x, \mu, a, c, p, q) &= (b\bar \mu a- c) p+  \frac{1}{2} \bar\mu^2\sigma^2 q + \log (c)-a^2.
\end{align*}
It then follows that
\begin{align*}
& \int_{\R\times\R_+} \exp\Big(\frac{1}{\gamma}H(t, x, \mu, a, c, p, q)\Big)da dc\\
=& \exp \Big(\frac{1}{2\gamma} \bar\mu^2  \sigma^2 q\Big) \int_{\mathbb{R}}\exp\Big( \frac{1}{\gamma} ab p \bar\mu -\frac{1}{\gamma}a^2 \Big)da\int_{\R_+} \exp\Big(\frac{1}{\gamma} \log (c)  -  \frac{p}{\gamma} c \Big)dc\\
=& \sqrt{\gamma \pi}\exp \Big(\frac{1}{2\gamma} \bar\mu^2  \sigma^2 q\Big) \exp\left( \frac{b^2p^2\bar{\mu}^2}{4\gamma}\right) \frac{\Gamma(1 + 1/\gamma)}{(\frac{p}{\gamma})^{1 + 1/\gamma}}\end{align*}
With the terminal condition $J^*(T, \mu) =0$, we can write the exploratory HJB equation by
\begin{align}
& \frac{\partial J^*}{\partial t}(t, \mu) -\beta J^*(t, \mu) + \gamma \int_{\R_+} \log \int_{\R \times \R_+} \exp\Big(\frac{1}{\gamma}H(t, {x}, \mu, a, c, \partial_\mu J^*(t, \mu)({x}), \partial_{x}\partial_\mu J^*(t, \mu)({x})\Big) da dc\mu(d{x}) \nonumber\\
= & \frac{\partial J^*}{\partial t}(t, \mu)  - \beta J^*(t, \mu) + \frac{1}{2} \bar\mu^2 \sigma^2 \int_{\R_+} \partial_{x}\partial_\mu J^*(t, \mu)({x}) \mu(d{x}) +\frac{1}{4}b^2\bar{\mu}^2  \int_{\mathbb{R}_+}(\partial_\mu J^*(t,\mu)({x}))^2\mu(d{x})\nonumber\\
&\; + \frac{\gamma}{2}\log(\gamma\pi)+ \gamma \log \Gamma(1+ \frac{1}{\gamma}) - (1 + \gamma) \int_{\R_+}\log \Big(\frac{\partial_\mu J^*(t, \mu)({x})}{\gamma}\Big)\mu(d{x}). \label{Ex_only_consumption_exploratory}
\end{align}
Suppose that $J^*(t, \mu)$ satisfies the following form
\begin{align*}
J^*(t, \mu) = B(t) \log \bar\mu +  D(t),
\end{align*}
with the terminal conditions $B(T)=D(T)=0$. It holds that
\begin{align*}
\frac{\partial J^*(t, \mu)}{\partial t} = \dot B(t) \log \bar\mu + \dot D(t),\quad  \partial_\mu J^*(t, \mu)(v) = B(t) \frac{1}{\bar\mu},\quad
\partial_v\partial_\mu J^*(t,\mu)(v) =0.
\end{align*}
Plugging these derivatives back into the exploratory HJB equation, we get that
\begin{align*}
&\left({\dot B(t)} -\beta B(t)+1+\gamma \right)\log \bar{\mu}+{\dot D(t)}-\beta D(t)+\frac{1}{4}b^2B^2(t)+\frac{\gamma}{2}\log(\gamma\pi) +\gamma\log\Gamma(1+1/\gamma)\\
&-(1+\gamma)\log\left(\frac{B(t)}{\gamma}\right)=0.
\end{align*}

Together with $B(T)=0$, we first get that $B(t)=\frac{1+\gamma}{\beta} (1-e^{-\beta(T-t)})$ and
\begin{align}\label{Dode}
&{\dot D(t)}-\beta D(t)+\frac{1}{4}b^2 \frac{(1+\gamma)^2}{\beta^2} (e^{-2\beta (T-t)} -2e^{-\beta(T-t)}+1) +K-(1+\gamma)\log(1-e^{-\beta (T-t)})=0,
\end{align}
where the constant $K:=\frac{\gamma}{2}\log(\gamma\pi) +\gamma\log\Gamma(1+1/\gamma)-(1+\gamma)\log\left(\frac{1+\gamma}{\beta\gamma}\right)$.

We can then obtain the explicit solution $D(t)$ of the ODE \eqref{Dode} with $D(T)=0$ as
\begin{align*}
D(t)=&A_1 e^{-2\beta(T-t)}+A_2e^{-\beta(T-t)}+A_3(1-e^{-\beta(T-t)})\log(1-e^{-\beta(T-t)})+A_4 t e^{-\beta(T-t)}+A_5,
\end{align*}
where
\begin{align*}
&A_1:=-\frac{b^2(1+\gamma)^2}{4\beta^3},\quad A_2:= (1+\gamma)T-\frac{b^2(1+\gamma)^2}{2\beta^2}T-\frac{K}{\beta},\\
&A_3:=-\frac{1+\gamma}{\beta},\quad A_4:=-(1+\gamma)+\frac{b^2(1+\gamma)^2}{2\beta^2},\quad A_5:=\frac{b^2(1+\gamma)^2}{4\beta^3}+\frac{K}{\beta}.
\end{align*}

As a result, we have the explicit essential q-function $q_e^*$ as
\begin{align*}
q_e^*(t, x, \mu, a, c)=&\frac{\partial J^*(t,\mu)}{\partial t}-\beta J^*(t,\mu)+ H(t, x, \mu, a, c, \partial_\mu J^*(t, \mu)(x), \partial_v\partial_\mu J^*(t, \mu)(x))\\
=&-a^2 + \frac{1+\gamma}{\beta}b(1-e^{-\beta(T-t)}) a -\frac{1+\gamma}{\beta \bar\mu}(1-e^{-\beta(T-t)})c +\log c -(1+\gamma)\log\bar{\mu}\\
&-\frac{b^2(1+\gamma)^2}{4\beta^2}(e^{-\beta (T-t)} -1)^2 -K + (1 + \gamma) \log (1 - e^{-\beta(T-t)}).
\end{align*}

Moreover, the separation form holds that ${\bm \pi}^*(a,c|t,x,\mu)={\bm \pi}_1^*(a|t,x,\mu){\bm \pi}_2^*(c|t,x,\mu)$, where ${\bm \pi}^*_1(a|t,x,\mu):=\mathcal{N}\Big( \frac{b(1+\gamma)}{2\beta} (1-e^{-\beta(T-t)}),\frac{\gamma}{2}\Big)$ and ${\bm \pi}_2^*(c|t,x,\mu):=\text{Gamma}(1+\frac{1}{\gamma}, \frac{1+\gamma}{\gamma \beta \bar\mu}(1-e^{-\beta(T-t)}))$.

Therefore, when the model parameters are unknown, we can parameterize the optimal value function $J$, the optimal essential q-function $q_e$ and the optimal policy ${\bm \pi}^*$ respectively that
\begin{align*}
&J^\theta(t, \mu) = \frac{1+\gamma}{\beta}(1-e^{-\beta(T-t)})\log\bar{\mu}+\theta_1 e^{-2\beta(T-t)}+\theta_2e^{-\beta(T-t)}\\
&-\frac{1+ \gamma}{\beta}(1-e^{-\beta(T-t)})\log(1-e^{-\beta(T-t)})+\theta_3 t e^{-\beta(T-t)}+\theta_4,\\
&q_e^\psi(t, x, \mu, a,c)=  -(1 + \gamma)\log\bar{\mu}+\psi_1(1-e^{-\beta(T-t)}) a -a^2 + \psi_2(e^{-\beta (T-t)} -1)^2\\
&-\frac{1+\gamma}{\beta \bar\mu}(1-e^{-\beta(T-t)}) c+\log c  -K + (1 + \gamma) \log (1 - e^{-\beta(T-t)})
 \\
&{\bm \pi}_1^\psi(a|t, x, \mu)  =\mathcal{N}\Big(\frac{\psi_1}{2} (1-e^{-\beta(T-t)}),\frac{\gamma}{2}\Big),\ {\bm \pi}_2^\psi(c|t, x, \mu) = \text{Gamma}(1+\frac{1}{\gamma}, \frac{1+\gamma}{\gamma \beta \bar\mu}(1-e^{-\beta(T-t)})).
\end{align*}
The parameters are $\theta_i$, $i=1,\ldots,4$ and $\psi_j$, $j=1,2$. We also require $q_2^\psi$ to satisfy \eqref{parametrized-coupled-value-constraint}. In this case, $\psi_2 =  -\frac{\psi_1^2}{4}$.

The following simulator will be used in Algorithms \ref{algo:offline episodic ml-method1}-\ref{algo:online episodic ml} to generate sample trajectories.

\noindent {\bf Simulator}\; Under the stochastic policy ${\bm h}^{\tilde\psi}= {\bm h}_1^{\tilde\psi} {\bm h}_2^{\tilde\psi}$, the dynamics \eqref{equ:mean-field-consumption-dynamics} becomes
\begin{align*}
d  \tilde X_s =& b\E[\tilde X_s]  \int_{\R} a {\bm h}_1^{\tilde\psi}(a|s, \tilde X_s, \P_{\tilde X_s}) da ds  - \int_{\R_+} c {\bm h}_2^{\tilde\psi}(c|s, \tilde X_s, \P_{\tilde X_s}) dc ds + \sigma \E[ \tilde X_s] dW_s\\
= & \frac{\tilde\psi_1 b}{2} (1-e^{-\beta(T-s)})\E[\tilde X_s]ds - \frac{\beta }{1 - e^{-\beta(T-s)}} \E[\tilde X_s]ds + \sigma \E[ \tilde X_s] dW_s.
\end{align*}
By taking the expectation on both sides of the above SDE, we obtain that
\begin{align*}
d \E[\tilde X_s] = \frac{\tilde\psi_1 b}{2} (1-e^{-\beta(T-s)})\E[\tilde X_s]ds - \frac{\beta }{1 - e^{-\beta(T-s)}} \E[\tilde X_s]ds.
\end{align*}
The Euler approximation of $\bar\mu_t = d \E[\tilde X_s]$ is
\begin{align}\label{R&Dlog-barmu-update}
\log \bar\mu_{t_{k+1}} = \log \bar\mu_{t_k} + \Big(\frac{\tilde\psi_1 b}{2} (1-e^{-\beta(T-s)}) - \frac{\beta }{1 - e^{-\beta(T-s)}}\Big) \Delta t.
\end{align}
The simulated reward is then given by
\begin{align*}
\hat r_{t_k} &= \int_{\R}\int_{\R_+} \log c {\bm h}_2^{\tilde\psi}(c|t_k, x, \P_{\tilde X_{t_k}}) dc \P_{\tilde X_{t_k}}(dx) - \int_{\R}\int_{\R} a^2 {\bm h}_1^{\tilde\psi}(a|t_k, x, \P_{\tilde X_{t_k}})da \P_{\tilde X_{t_k}}(dx)\\
& = \frac{\Gamma'(1 + \frac{1}{\gamma})}{\Gamma(1 + \frac{1}{\gamma})} - \log\big(\frac{1+\gamma}{\gamma \beta}(1-e^{-\beta(T-t)})\big) + \log \bar\mu +  \frac{\tilde\psi_1^2}{4} (1-e^{-\beta(T-t)})^2 + \frac{\gamma^2}{4}.
\end{align*}

We first set coefficients of the simulator to $T = 1, b =0.5, \sigma = 0.5$. We next set the known parameters as: $\gamma = 0.25$, $\beta = 2$, the time step $\Delta t = 0.1$, the number of episodes $N = 8000$, the number of test policies $M = 10$,  the lower bound of uniform distribution $p(j) = 0$, the upper bound of uniform distribution $q(j) = \frac{2}{j^{0.6}}$, the log mean $\log \mu_{0}^m, 1 \leq m \leq M$ that is initialed to be 0, and the learning rates $(\alpha_\theta, \alpha_\psi)$ are chosen by
\[
 \alpha_\theta(j) =
 \left\{
 \begin{array}{lll}
 (\frac{0.05}{j^{0.22}}, \frac{0.2}{j^{0.15}}, \frac{0.2}{j^{0.25}}, 0.5), \; & \mbox {if}\; j \leq 4500, \\
 (\frac{0.05}{j^{0.4}}, \frac{0.2}{j^{0.15}}, \frac{0.2}{j^{0.55}}, \frac{0.5}{j^{0.01}}), \; & \mbox {if}\; 4500 < j \leq 8000,
 \end{array}
 \right.
 \]
 and
 \begin{align*}
 \alpha_\psi(j) =
 \left\{
 \begin{array}{lll}
 \frac{0.15}{j^{0.31}}, \; & \mbox {if}\; j \leq 4500, \\
\frac{0.15}{j^{0.61}}, \; & \mbox {if}\; 4500 < j \leq 8000.
 \end{array}
 \right.
 \end{align*}

\begin{figure}[tbhp]
\centering
\includegraphics[width=0.4\linewidth]{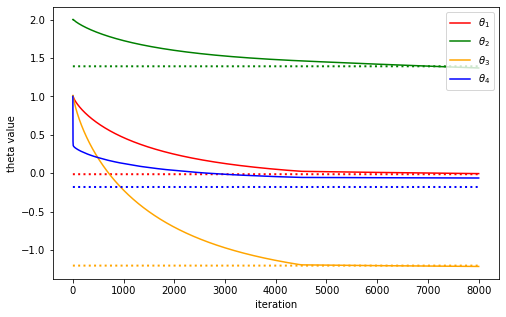}
\includegraphics[width=0.4\linewidth]{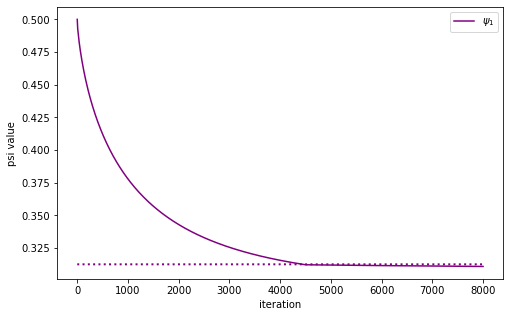}
\includegraphics[width=0.4\linewidth]{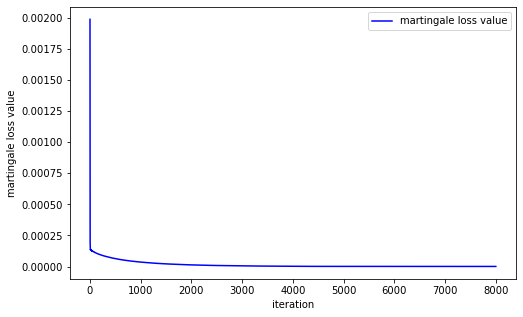}
\includegraphics[width=0.4\linewidth]{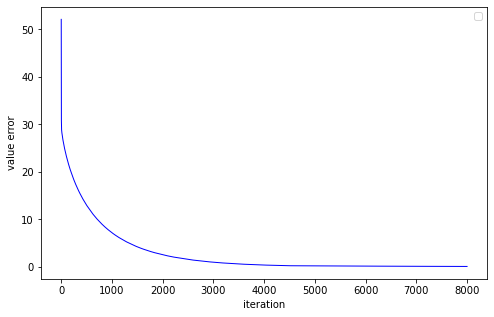}
\caption{\footnotesize {\bf Convergence of value function and q-function for Algorithm \ref{algo:offline episodic ml-method1}}. Paths of learnt parameters for value function (top left) and path of learnt parameters for q-function (top right) vs optimal parameters shown in the dashed line. The change of the weak martingale loss value over iterations (bottom left) and the change of $L^2$-error over iterations (bottom right) along a trajectory $(\log (\bar\mu_{t_k}))_k$ with $\log(\bar\mu_0) =0$ controlled by the learnt policy and by the optimal policy.}
\label{fig:algorithm1-conusmption}
\end{figure}

 Based on the offline q-learning Algorithm \ref{algo:offline episodic ml-method1}, we plot in Figure \ref{fig:algorithm1-conusmption} the numerical results on the convergence of parameters $\theta$ and $\psi$ for the value function and q-function, and also summarize the learnt parameters for the optimal value function and the optimal essential q-function $q_e$ in Table \ref{table:parameters}.

 \begin{table}
 \caption{}
 \centering
 \begin{tabular}{cccccc}
 \hline
 & $\theta_1$ & $\theta_2$ & $\theta_3$ & $\theta_4$ & $\psi_1$ \\
 \hline
 true value &-9.765 $\times 10^{-5}$ & 1.085 & -1.248 & 0.163 & 0.0625\\
 \hline
 learnt by Algorithm \ref{algo:offline episodic ml-method1}   & 0.0796  & 1.107 & -1.281&0.119& 0.0667 \\
 \hline
 \end{tabular}
 \label{table:parameters}
 \end{table}

\section{Conclusion}\label{sec:conclusion}
This paper aims to lay the theoretical foundation of the continuous time q-learning for mean-field control problems, which can be viewed as the bridge between the discrete time Q-learning with the integrated Q-function for MFC problems studied in \cite{GGWX23} and the continuous time q-learning with the q-function for single agent's control problems studied in \cite{jiazhou2022}. In our framework, two different q-functions are introduced for the purpose of learning, namely the integrated q-function $q$ that is defined as the first order derivative of the integrated Q-function with respect to time and the essential q-function $q_e$ that is used for policy improvement iterations. Comparing with two counterparts in \cite{GGWX23} and \cite{jiazhou2022}, we establish the weak martingale characterization of the value function and the integrated q-function through test polices in the neighbourhood of the target policy in a similar flavor of off-policy learning. To learn the essential q-function from the double integral representation, we propose the average-based loss function by
searching test policies in the same form of the parameterized target policy but with randomized parameters. In two examples, one in the LQ control framework and one beyond the LQ control framework, we can illustrate the effectiveness of our q-learning algorithm.

Several interesting future extensions can be considered. Firstly, we may consider the controlled common noise in the mean-field control problem, where the optimal policy no longer admits the explicit form as a Gibbs measure. Instead, we can provide the first order condition of the optimal policy using the linear functional derivative with respect to probability measures. It is an interesting open problem to investigate the correct form of the continuous time integrated q-function and the associated q-learning. Secondly, it will also be appealing to investigate the decentralized continuous time q-learning from the representative agent's perspective and provide the correct q-function and policy iteration rules based on observations of the representative agent's individual state dynamics. At last, we are also interested in establishing some theoretical convergence results on the policy improvement iterations and continuous time q-learning in the mean field model.

\ \\
\noindent
\textbf{Acknowledgement}:  We sincerely thank two referees for their constructive comments and suggestions. X. Wei is supported by National Natural Science Foundation of China grant under no.12201343.  X. Yu is supported by the Hong Kong Polytechnic University research grant under no. P0045654.

\section*{Declarations}

 \textbf{Competing interests} The authors have no competing interests to declare that are relevant to the content of this article.

 \appendix
 \section{Connection between formulations \eqref{equ:exploratory_SDE} and \eqref{equ:exploratory_average_SDE}} \label{appenA}
 \begin{Lemma}
 ${X_s^{{\bm \pi}}}$  and $\tilde X_s^{\bm \pi}$, as solutions to \eqref{equ:exploratory_SDE} and \eqref{equ:exploratory_average_SDE} respectively, have the same distribution for each $s \in [t, T]$ as they correspond to the same martingale problem.
 \end{Lemma}
{\bf Proof}.\; Let $\Cc_c^\infty(\R^d)$ denote the set of infinitely differentiable function $\varphi: \R^d \to \R$ with compact set, and let $\nabla \phi$ and $\nabla^2\phi$ denote the gradient and Hessian of $\varphi$, respectively. We define the infinitesimal generator
  \begin{align} \label{generatorL}
  L_s^a \varphi := b(s, x,  \mu, a)\trans \nabla \varphi(x) + \frac{1}{2} {\rm Tr}\big(\sigma\sigma\trans(s, x, \mu, a) \nabla^2\varphi\big).
  \end{align}
  Given $(t, \mu) \in [0, T] \times \Pc_2(\R^d)$ and fixed ${\bm \pi} \in \Pi$, a solution to the martingale problem for the generator $L^a$ in \eqref{generatorL} is a probability measure $m$ such that
  \begin{align}\label{martingale-formulation}
\varphi(X_s) - \int_t^s \int_{\Ac} L_u^a \varphi(u, X_u, m_{X_u}, a) {\bm \pi}(a|u, X_u, m_{X_u})da du,\;\; s \geq t
\end{align}
is a $m$-martingale for all $\varphi \in \Cc_c^\infty(\R^d)$. The martingale formulation \eqref{martingale-formulation} was studied in \cite{KM1990} and then generalized to MFC theory in \cite{lacker2017} for the limit theory of $N$-player dynamics to the controlled McKean-Vlasov dynamics.

Note that when $\tilde X_s^{\bm \pi}$ satisfies \eqref{equ:exploratory_average_SDE}, It\^o's formula implies that $\tilde X^{\bm \pi}$ satisfies \eqref{martingale-formulation}. Furthermore, we can also apply It\^o's formula to $X_s^{\bm \pi}$ satisfying \eqref{equ:exploratory_SDE} and take the conditional expectation given $\Fc_r$, $t \leq r \leq s$,
\begin{align*}
&\E[\varphi(X^{\bm \pi}_s)|\Fc_r] -  \varphi(\xi)
 = \E\Big[\int_t^s L_u^{a^{\bm \pi}}\varphi(u, X_u^{\bm \pi}, \P_{X_u}^{\bm \pi}) du \big|\Fc_r \Big]\\
 =& \E\Big[\int_t^s L_u^{a}\varphi(u, X_u^{\bm \pi}, \P_{X_u}^{\bm \pi}) {\bm \pi}(a|u, X_u, \P_{X_u})du\big|\Fc_r\Big],
\end{align*}
where in the last equality we have used the fact the action $a^{\bm \pi}$ is sampled from ${\bm \pi}$ independent of $W$, and hence $\E[\varphi(X_u, \P_{X_u}, a_u)|\Fc_r] = \E[\int_{\Ac}\varphi(X_u, \P_{X_u}, a){\bm \pi}(a|u, X_u, \P_{X_u})da|\Fc_r]$ for all $\varphi$.
Therefore, $X_s^{\bm \pi}$ also satisfies \eqref{martingale-formulation}. It follows from the uniqueness of the martingale problem that $\tilde X_s^{\bm \pi}$ and $X_s^{\bm \pi}$ are the same in law.

\end{document}